\documentclass[10pt,twocolumn,letterpaper]{article}

\usepackage[pagenumbers]{cvpr} % To force page numbers, e.g. for an arXiv version

\usepackage{graphicx}
\usepackage{times}
\usepackage{helvet}
\usepackage{courier}
\usepackage{amsmath}
\usepackage{algorithm}
\usepackage{algorithmic}
\usepackage{csquotes} 
\usepackage{color}
\usepackage{paralist}
\usepackage{amssymb}
\usepackage{indentfirst}
\usepackage{float}
\usepackage{multirow}
\usepackage{cite}
\usepackage{mathrsfs}

\usepackage{mathrsfs}
\usepackage{textcomp,booktabs}
\usepackage{amssymb}% http://ctan.org/pkg/amssymb
\usepackage{pifont}% http://ctan.org/pkg/pifont
\usepackage[misc]{ifsym}

\usepackage{mathrsfs}
\usepackage{color}
\usepackage{soul}
\usepackage{colortbl}
\usepackage[dvipsnames]{xcolor}
\definecolor{bleu1}{named}{SkyBlue}
\definecolor{bleu2}{named}{CornflowerBlue}
\definecolor{bleu3}{named}{Cerulean}
\definecolor{bleu4}{named}{RoyalBlue}
\definecolor{rouge}{named}{Red}
\definecolor{meteor}{named}{ForestGreen}
\definecolor{cider}{named}{Orange}

\definecolor{citecolor}{HTML}{0071bc} 
\definecolor{SeaGreen4}{RGB}{0,205,102} 
\definecolor{SlateBlue}{RGB}{106,90,205} 
\definecolor{DarkRed}{RGB}{178,34,34} 
\usepackage[colorlinks, linkcolor=red,  anchorcolor=blue, citecolor=citecolor]{hyperref}

\usepackage{pgfplotstable} 
\usepackage{pgfplots}
\pgfplotsset{compat=newest} % 使用最新版本的pgfplots
\usepackage{pgffor} % 循环宏包
\usepackage{filecontents}
\usepackage{enumitem}

\usepackage{colortbl}
\definecolor{mygray}{gray}{.9}
\definecolor{mypink}{rgb}{.99,.91,.95}
\definecolor{mycyan}{cmyk}{.3,0,0,0}
\newcommand{\modification}[1]{#1}                
\newcommand{\mmodification}[1]{#1}                
 \newcommand{\mmmodification}[1]{{\color{black}#1}}                

\newcommand{\tabref}{Table}  
\usepackage{color}
\definecolor{citecolor}{HTML}{0071bc} 
\definecolor{SeaGreen4}{RGB}{0,205,102} 
\definecolor{SlateBlue}{RGB}{106,90,205} 
\definecolor{DarkRed}{RGB}{178,34,34} 

\usepackage[capitalize]{cleveref}
\crefname{section}{Sec.}{Secs.}
\Crefname{section}{Section}{Sections}
\Crefname{table}{Table}{Tables}
\crefname{table}{Tab.}{Tabs.}

\definecolor{cvprblue}{rgb}{0.21,0.49,0.74}

\def\paperID{9748}      % *** Enter the Paper ID here

\def\confName{CVPR}
\def\confYear{2024}

\title{R2GenCSR: Mining Contextual and Residual Information for LLMs-based Radiology Report Generation}

\author{Xiao Wang$^{1}$, Yuehang Li$^{1}$, Fuling Wang$^{1}$, Shiao Wang$^{1}$, 
        Chuanfu Li$^{2}$, Bo Jiang$^{1}$\thanks{Corresponding Author: Bo Jiang} \\ 
${^1}$ {School of Computer Science and Technology, Anhui University, Hefei, China} \\
${^2}$ {First Affiliated Hospital of Anhui University of Chinese Medicine, Hefei, China} \\  
\textit{xiaowang@ahu.edu.cn}, 
\textit{e23201112@stu.ahu.edu.cn}, 
\textit{18870722722@163.com}, \\ 
\textit{wsa1943230570@126.com}, 
\textit{licf@ahtcm.edu.cn}, 
\textit{zeyiabc@163.com} 
}

\begin{document}
\maketitle

%%%%%%%%% ABSTRACT
\begin{abstract}
Inspired by the tremendous success of Large Language Models (LLMs), existing Radiology report generation methods attempt to leverage large models to achieve better performance. They usually adopt a Transformer to extract the visual features of a given X-ray image, and then, feed them into the LLM for text generation. How to extract more effective information for the LLMs to help them improve final results is an urgent problem that needs to be solved. Additionally, the use of visual Transformer models also brings high computational complexity. To address these issues, this paper proposes a novel context-guided efficient radiology report generation framework. Specifically, we introduce the Mamba as the vision backbone with linear complexity, and the performance obtained is comparable to that of the strong Transformer model. More importantly, we perform context retrieval from the training set for samples within each mini-batch during the training phase, utilizing both positively and negatively related samples to enhance feature representation and discriminative learning. Subsequently, we feed the vision tokens, context information, and prompt statements to invoke the LLM for generating high-quality medical reports. Extensive experiments on three X-ray report generation datasets (i.e.,  \mmodification{IU X-Ray}, MIMIC-CXR, CheXpert Plus) fully validated the effectiveness of our proposed model. 
The source code is available at \url{https://github.com/Event-AHU/Medical_Image_Analysis}.
% \textcolor{red}{The source code of this work will be released upon acceptance.} 
\end{abstract}

\section{Introduction}
\label{sec:introduction}
X-ray image based radiology report generation is one of the typical applications of Artificial Intelligence (AI) in healthcare. It aims to utilize powerful AI models to directly generate high-quality medical reports from given X-ray images, thereby alleviating the workload of doctors and reducing patient waiting times. Although the performance of this task has made significant strides with the development of AI, it still falls short of matching the expertise of professional physicians due to various challenges. Due to privacy concerns surrounding X-ray data, there is a lack of quality and diversity in the training datasets, and the rarity of certain diseases and abnormal conditions also results in poor generalization performance of the models. Therefore, the research on X-ray image based report generation is still a very important but unsolved problem.

\begin{figure}
 \centering
 \includegraphics[width=1\linewidth]{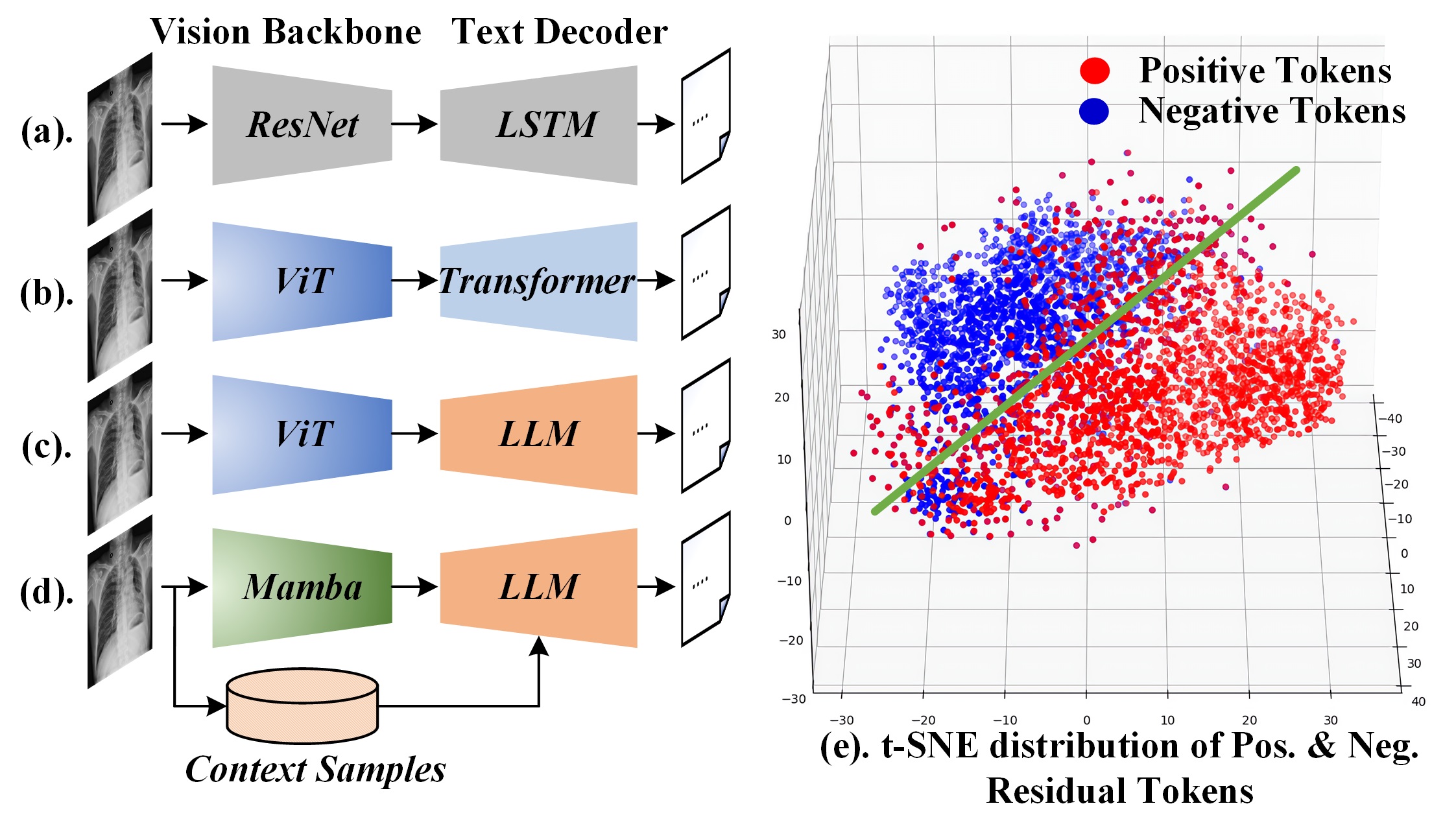}
 \caption{
    \mmodification{Comparison between (a-c). existing X-ray report generation frameworks and (d). our newly proposed one; 
    (e). t-SNE feature distribution of our sampled positive and negative context samples from \mmodification{IU X-Ray} dataset.} } 
 \label{fristIMG}
\end{figure}

Due to the good generalization of deep learning, the performance of radiology report generation has seen steady improvements. For example, Cao et al. propose the Multi-modal Memory Transformer Network (MMTN)~\cite{cao2023mmtn} for image-report consistent radiology report generation. Jin et al. introduce the PromptMRG~\cite{jin2024promptmrg} which attempts to enhance the diagnostic accuracy in the report by feeding diagnosis-aware prompts. Li et al. proposed a new radiological reports DCL~\cite{li2023dynamic} framework that uses dynamic graphs to integrate specific and general knowledge to improve visual representation. This framework also enhances visual and textual representation by leveraging contrastive learning objectives. KiUT~\cite{huang2023kiut} utilizes the U connection between the encoder and decoder to effectively integrate different levels of visual information. It introduces a knowledge graph and uses a distillation technique to produce reports that are more aligned with real-world conditions. METransformer~\cite{Wang_2023_CVPR} simulates multi-expert joint diagnostics by introducing multiple learnable ``expert" tokens. These tokens focus on different areas while interacting with each other to capture reliable and complementary visual information, facilitating the parallel generation of diagnostic reports.
Although these models have achieved good results, their overall performance is still far from reaching the level of human experts. How to narrow the gap further with professional doctors remains a question worth pondering and researching.

Inspired by the great success of LLMs in natural language processing (NLP), some researchers also introduced LLMs for radiology report generation. Specifically, Liu et al.~\cite{liu2024bootstrapping} propose to augment the LLMs-based radiology report generation using in-domain instance induction and coarse-to-fine decoding. 
Although better performance can be obtained, these models still achieve their efficiency and performance bottlenecks due to the following issues: 
\textbf{Firstly}, the performance of current LLMs heavily relies on the tokens humans feed into, for example, the prompt sentence, visual tokens, etc. More comprehensive inputs can guide the LLMs to generate high-quality medical reports. However, seldom of current works consider the context samples (e.g., the samples with/without disease) which may be very important cues for the text generation of the current sample. 
\textbf{Secondly}, they adopt the Transformer network as the vision backbone which is computationally expensive ($\mathcal{O}(N^2)$). When handling long-range visual tokens (e.g., the raw X-ray images are usually high-definition), the self-attention in the Transformer often performs poorly on the speed, memory usage, etc. 

With these issues in mind, in this work, we propose a novel radiology report generation framework that adopts the Mamba~\cite{gu2023mamba} as the backbone and mines the context samples to guide the large language models for high-quality report generation. A comparison between existing models and ours is illustrated in Fig.~\ref{fristIMG} (a-d). 
We partition the input X-ray image into patches and project them into visual tokens. Then, the Mamba backbone network is adopted for efficient and effective visual token processing. More importantly, we retrieve some context samples from the training subset for each image in the mini-batch to help our report generator understand which samples are likely to have diseases and which do not. \modification{The motivation for retrieving both positive and negative samples is to provide the LLM with contrastive contextual cues: positive cases highlight disease-related visual patterns, while negative cases serve as normal references. This contrast helps the model to better capture the subtle residual differences between abnormal and normal structures, thereby guiding the LLM to generate more accurate and clinically meaningful reports.} These context samples are also processed using the Mamba backbone and subtracting the global tokens of the current image to get the residual tokens. We provide the context prompts to assist the LLM in distinguishing whether certain tokens are related to diseases. As shown in Fig.~\ref{fristIMG} (e), it is easy to find that these retrieved positive and negative samples are easily distinguishable from the perspective of the t-SNE~\cite{van2008tSNE} feature distribution. Finally, we feed these visual tokens, context tokens, and prompts into the LLM for high-quality radiology report generation. Extensive experiments on two widely used benchmark datasets fully validated the effectiveness of our proposed framework.

To sum up, the main contributions of this work can be listed as follows: 

1). We propose a novel large language model based radiology report generation framework that is augmented by context samples in the training phase, termed R2GenCSR. Our approach effectively utilizes both positive and negative samples to guide the model, enhancing its ability to generate more accurate and contextually relevant reports.

% 2). We propose an efficient and effective half-precision vision Mamba that achieves comparable performance to the widely used Transformer backbone network for the radiology report generation task. 
% 2). We introduce a residual-guided approach for improved report generation by capturing the semantic differences between visual and textual information, providing  a new perspective on combining multi-modal data (medical images and clinical texts) without explicit alignment. 
2). We introduce a residual-guided approach for improved report generation by capturing the semantic differences between visual and textual information, providing a new perspective on combining multi-modal data (medical images and clinical texts).

3). Extensive experiments on the widely used \mmodification{IU X-Ray}, MIMIC-CXR, and CheXpert Plus datasets fully validated the effectiveness of our proposed X-ray report generation framework.  
% 3). Extensive experiments on the widely used \mmodification{IU X-Ray}, MIMIC-CXR, and CheXpert Plus datasets were conducted to evaluate the performance of our proposed radiology report generation framework, demonstrating superior performance in terms of both accuracy compared to existing methods.

\section{Related Work}

\subsection{Radiology Report Generation} 
Radiology report generation (RRG), also referred to as medical report generation (MRG), focuses on generating diagnostic reports from medical images. Existing approaches can be categorized into CNN-based, RNN-based, and Transformer-based frameworks. To be specific, Li et al.~\cite{li2020CRNN} propose a model that combines CNNs and RNNs to generate medical reports from chest X-ray images. Jing et al.~\cite{Jing_2018} propose to predict the possible diseases using LSTM~\cite{Hochreiter1997LSTM} and then generate medical reports based on those predictions. Chen et al.~\cite{chen2020R2Gen} demonstrate the effectiveness of generating detailed and accurate radiological reports from X-ray images using the Transformer model, leveraging visual and text data to improve performance. Yang et al.~\cite{yang2021joint} addresses the visual-to-semantic gap in medical image report generation by employing a triple-branch network to jointly encode deep visual and semantic embeddings. Huang et al.~\cite{KCAP} leverages a medical knowledge dictionary to bridge the semantic gap between visual and text modalities. Wang et al.~\cite{wang2024XrayHDMAE} pre-train a ViT model on the high-resolution X-ray images using masked auto-encoder for radiology report generation. Zhang et al.~\cite{MPMA} proposes a unified Med-VLP framework with multi-task paired masking with alignment module to enhance cross-modal interaction and semantic representation. Different from existing works, in this paper, we propose a novel context sample retrieval-guided LLMs framework for radiology report generation, thus, our model becomes more sensitive and accurate in predicting diseases.

\subsection{Large Language Models} 

The integration of Large Language Models (LLMs) into radiology report generation has become a prominent research direction. LLMs can be categorized into two types: base LLMs and domain-specific LLMs. Early foundational models like Google's BERT~\cite{devlin2019bert} and Meta's LLaMA~\cite{touvron2023llama} have demonstrated significant advancements in natural language processing. OpenAI's GPT series, particularly GPT-3~\cite{brown2020language}, has further pushed the boundaries of language generation and understanding with its 175 billion parameters.

In the medical domain, LLMs have been tailored for specific tasks. R2Gen-GPT~\cite{R2GenGPT} integrates image and text features into the LLaMA2-7B~\cite{touvron2023llama2} decoder for report generation. RGRG~\cite{Tanida2023RGRG} adopts a GPT-2~\cite{radford2019language}-based approach to generate and reconnect sentences for detected regions. Google's Med-PaLM2~\cite{singhal2023Med-PaLM2}, built on PaLM 2~\cite{anil2023palm2}, achieve state-of-the-art performance through domain-specific fine-tuning. MedVersa~\cite{zhou2024MedVersa} extends capabilities to multimodal medical inputs and real-time task specification. Inspired by these works, we propose enhancements to radiology report generation using LLMs.

\subsection{State Space Model} 
To address the high computational cost of Transformer networks, State Space Models (SSMs)~\cite{wang2024SSMSurvey} have been proposed to achieve linear complexity. The S4 model~\cite{S4} introduces a structured state space approach, improving long-sequence dependency modeling while enhancing efficiency and accuracy. S5~\cite{smith2023simplified} further simplifies and optimizes S4, balancing computational efficiency and sequence modeling capabilities. Mamba~\cite{gu2023mamba} advances SSMs with a time-varying state-space model and selection mechanism, enabling efficient long-sequence modeling.In vision-related tasks, SSMs have shown promising results. Vim~\cite{zhu2024visionMamba} (Vision Mamba) achieves high performance and efficiency by incorporating positional embeddings and bidirectional SSMs for high-resolution image processing. VMamba~\cite{liu2024vmamba} introduces a Cross-Scan Module (CSM) to ensure global receptive fields with linear complexity. Mamba-2~\cite{Dao2024mamba2} further enhances the Mamba architecture by integrating SSD theory and structural attention mechanisms.

Inspired by the linear computational cost, in this work, we propose to encode the X-ray image using a half-precision vision Mamba network and achieve similar performance on three radiology report generation benchmark datasets.

\subsection{Context Sample Retrieving}  
Retrieval-Augmented Generation (RAG) is a hybrid approach that combines retrieval and generation to enhance the performance of NLP tasks, particularly those requiring extensive knowledge and contextual information. Recent advancements in retrieval-augmented methods have demonstrated their effectiveness across various domains.
%%%% 
For instance, MedGraphRAG~\cite{wu2024medicalGraphRAG} proposes a graph-based U-Retrieval framework that combines Top-down Precise Retrieval with Bottom-up Response Refinement to balance global context awareness with fine-grained indexing. Similarly, BGFormer~\cite{wang2024BGFormer} introduces a Structure-constrained Self-attention (SSA) mechanism to mine relationships between samples for robust data representation, while CricaVPR~\cite{Lu_2024_CVPR} leverages cross-image differences like perspective and illumination to generate more robust image features. CRAG~\cite{yan2024CRAG} enhances retrieval quality through a search evaluator and large-scale web searches, and EgoInstructor~\cite{Xu_2024_EgoInstructor} utilizes third-person video resources to improve first-person video understanding. Additionally, RALF~\cite{Horita_2024_RALT} retrieves similar layout examples to address challenges in capturing high-dimensional layout structures, and EVCAP~\cite{Li_2024_EVCap} constructs external visual-name memory for retrieval-augmented image captioning. \modification{Contrastive Attention (CA)~\cite{liu2021contrastive} leverages differences between positive and negative X-ray samples to highlight disease-specific features. In contrast, our residuals are computed in the shared embedding space of visual features and textual prompts, guiding the LLM to better distinguish disease-relevant information for more accurate report generation.}

In the field of medical report generation, RAG can also serve as a clinical decision support tool by combining medical databases and research papers, helping physicians quickly access the latest research on disease diagnosis, treatment options, and drug information. Our extensive experiments on three benchmark datasets support the effectiveness of context samples for the medical report generation task.

\begin{figure*}
\centering
\includegraphics[width=1\linewidth]{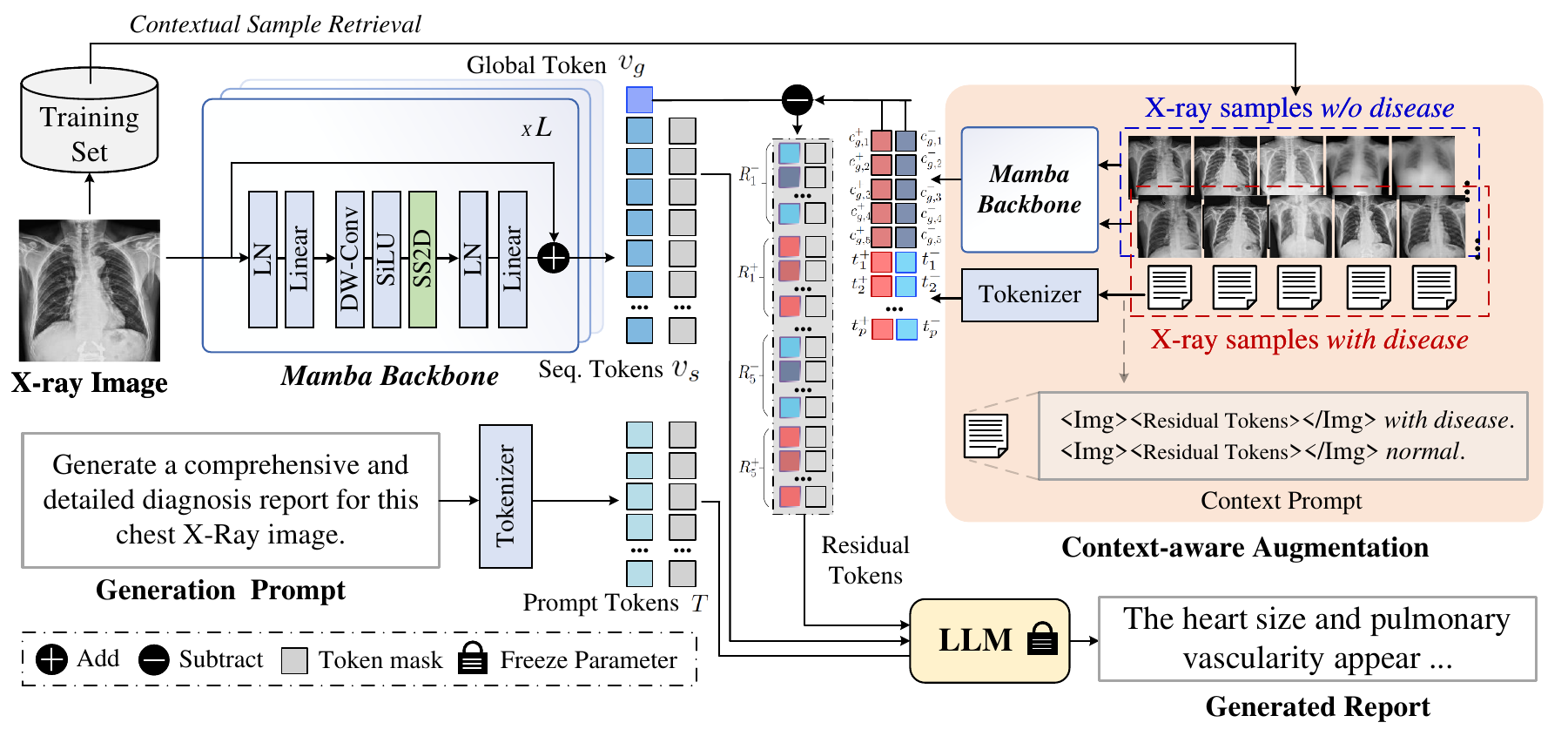}
\caption{\modification{An overview of our proposed context sample augmented large language model for efficient radiology report generation, termed R2GenCSR. Three main modules are involved in this framework, including the Mamba vision backbone, context sample retrieval, and large language model (LLM). We first extract the visual tokens of the input X-ray image using the Mamba backbone, then, retrieve context samples from the training subset. Residual tokens are then computed by subtracting the projected context sample tokens from the projected input image tokens, all within the LLM's text embedding space. The LLM takes the vision tokens, context residual tokens, and prompt statements as input to generate a high-quality medical report. }} 
\label{fig:framework} 
\end{figure*}

\section{Methodology}  

In this section, we will first give a review of the Mamba network and an overview of our proposed R2GenCSR framework. Then, we will dive into the details of the R2GenCSR framework, with a focus on Input Representation, Context Sample Retrieval, LLM for Report Generation, and Loss Function. More details will be introduced in the subsequent subsections.

\subsection{Preliminary: Mamba}  
\mmmodification{
Current widely used Mamba networks are developed based on the continuous SSMs. It maps a one-dimensional function or sequence 
\( x(t) \in \mathbb{R}^p \) to \( y(t) \in \mathbb{R}^q \) 
through a hidden state 
\( h(t) \in \mathbb{R}^n \). 
The computing procedure can be summarized as follows: 
\begin{align}
    h'(t) = \mathbf{A} h(t) + \mathbf{B} x(t), \quad
    y(t) = \mathbf{C} h(t), 
\end{align}
where $\mathbf{A} \in \mathbb{R}^{n \times n}$, $\mathbf{B} \in \mathbb{R}^{n \times p}$, $\mathbf{C} \in \mathbb{R}^{q \times n}$ denotes the state matrix, input matrix, and output matrix.

As the image and text we processed are discrete data, the aforementioned continuous SSMs needed to be transformed into discrete ones. For example, the S4~\cite{S4} and Mamba model adopts the Zero-Order Hold (ZOH) to realize this goal, i.e., 
\begin{align}
    \bar{\mathbf{A}} &= \exp(\Delta \mathbf{A}), \\
    \bar{\mathbf{B}} &= (\Delta \mathbf{A})^{-1} (\exp(\Delta \mathbf{A}) - \mathbf{I}) \cdot \Delta \mathbf{B},
\end{align}
where the \( \Delta \) is a timescale parameter (also called step size). Thus, we can reformulate the discrete version of SSMs as: 
\begin{align}
    h_t = \bar{\mathbf{A}} h_{t-1} + \bar{\mathbf{B}} x_t, \quad
    y_t = \mathbf{C} h_t.
    \label{eq:mamba}
\end{align}

% And also speed the training and inference using a couple of hardware-aware algorithms.
To further strengthen the SSM, Gu et al. propose the Mamba~\cite{gu2023mamba} which makes the model varying from time-invariant to dependent. Inspired by the success of Mamba in NLP, researchers also adapt it to the computer vision community such as  VMamaba~\cite{liu2024vmamba} and Vim~\cite{zhu2024visionMamba}. As shown in Eq.~\ref{eq:mamba}, the recursive structure offers two architectural advantages for X-ray images. First, the hidden state \(h_t\) serves as a cumulative memory of previously processed patches. When an image is flattened into patches, each new patch is integrated with the aggregated representation of all preceding patches. This creates an implicit global receptive field without the quadratic cost of explicit attention, which is clinically valuable }\mmmodification{for chest X-ray images where pathologies often manifest as spatially distributed patterns. Second, the selective mechanism in Mamba makes matrices \(\bar{\mathbf{A}}\) and \(\bar{\mathbf{B}}\) input-dependent, allowing dynamic adjustment of information retention. Patches containing abnormalities can be selectively emphasized in the state update, while normal anatomical structures can be compressed.

}

\subsection{Overview} 
In this paper, we propose a novel contextual sample retrieval guided large language model framework for efficient radiology report generation. As shown in Fig.~\ref{fig:framework}, it can be divided into three main parts, i.e., the Mamba vision backbone, context retrieval module, and large language model (LLM) for report generation. Given the X-ray image, we first extract its visual tokens using the Mamba backbone. Meanwhile, we retrieve context samples (X-ray samples with and without disease) from the training subset based on the input image and embed them into visual and text tokens. Then, the residual tokens which measure the difference between the input and context samples can be obtained via the subtract operator. Finally, we feed the vision tokens, context residual tokens, and prompt statements into the LLM to generate a high-quality medical report. One can note that the proposed framework stands out from existing methods by \textit{incorporating context retrieval} and \textit{using a linear complexity vision backbone}, which enhances feature representation and discriminative learning while maintaining computational efficiency.

\subsection{R2GenCSR Framework}   

In this subsection, we will introduce the R2GenCSR framework from the perspective of \textit{Input Representation}, \textit{Context Sample Retrieval}, \textit{LLM for Report Generation}, and \textit{Loss Function}.

% \noindent $\bullet$ 
\subsubsection{Input Representation} 
Assume the dataset contains $\mathbf{N}$ X-ray images, $\mathbf{X} =\{x_{1}, x_{2}, \dotsc, x_{N} \}$, where each $x_{i} \in \mathbb{R}^{C\times H\times W}$ represents a X-ray image with $C$ channels, height $H$, and width $W$. For each X-ray image $x$, the corresponding feature map $v \in \mathbb{R}^{\hat{H} \times \hat{W} \times \hat{C}} = VMamba(x)$ can be obtained after feeding it into the VMamba backbone, where $\hat{H}$ and $\hat{W}$ are the spatial dimensions of the feature map, and $\hat{C}$ is the number of feature channels. The reason our framework adopts VMamba instead of conventional visual Transformer models, such as ViT and Swin-Transformer, is because the computational complexity of this model is linear ($\mathcal{O}(N)$), requiring lower computational resources. As shown in Fig.~\ref{fig:framework}, the basic VMamba block consists of Layer Normalization (LN), Linear Layer, DW-Conv layer, SiLU activation layer, SS2D module, and also the skip connections. 
% \mmmodification{Unlike Swin Transformer that restricts self-attention to local windows and approximates global context through shifted-window operations, Mamba processes the entire sequence of image patches in a single recurrent pass with strict $\mathcal{O}(N)$ complexity. As shown in Eq.~\ref{eq:mamba}, the input-dependent projections $\bar{B}(x_t)$ and $C(x_t)$ enable content-aware selection of historical information from $h_{t-1}$. This mechanism allows VMamba to capture long-range anatomical dependencies (e.g., bilateral opacities in chest X-rays) without the quadratic cost or locality constraints of attention-based backbones.}

Then, two distinct types of representations are generated based on feature map $v$, i.e., global features $v_g$ and sequential tokens $v_s$. Specifically, a 2D global average pooling operation is applied to $v$ along the spatial dimensions, followed by a projection layer to obtain the global features $v_g$. The sequential tokens $v_s$ are derived by flattening $v$ along the spatial dimension and processing it through the same projection layer. The projection layer (Proj) is a linear layer \modification{without any activation function}, serves as an interface between the visual and linguistic modalities by mapping visual features to a dimensionality $E$ that aligns with the token embedding size of the language model. This operation yields global features $v_g \in \mathbb{R}^{E}$ and sequential tokens $v_s \in \mathbb{R}^{L \times E}$, both of which are compatible with the language model's processing requirements.  
Mathematically speaking, 
\begin{align}
v_g & = Proj(avgPool2D(v))\label{eq:v_g},\\
v_{s} & = Proj(Flatten(v)),
\end{align} 
Here, $v_g$ captures the global information of the X-ray image, while the sequential tokens $v_s$ represent local visual features. In our framework, $v_g$ is employed for residual computations due to its capacity to preserve compact yet comprehensive global representations. 
\begin{figure}
    \centering
    \includegraphics[width=\linewidth]{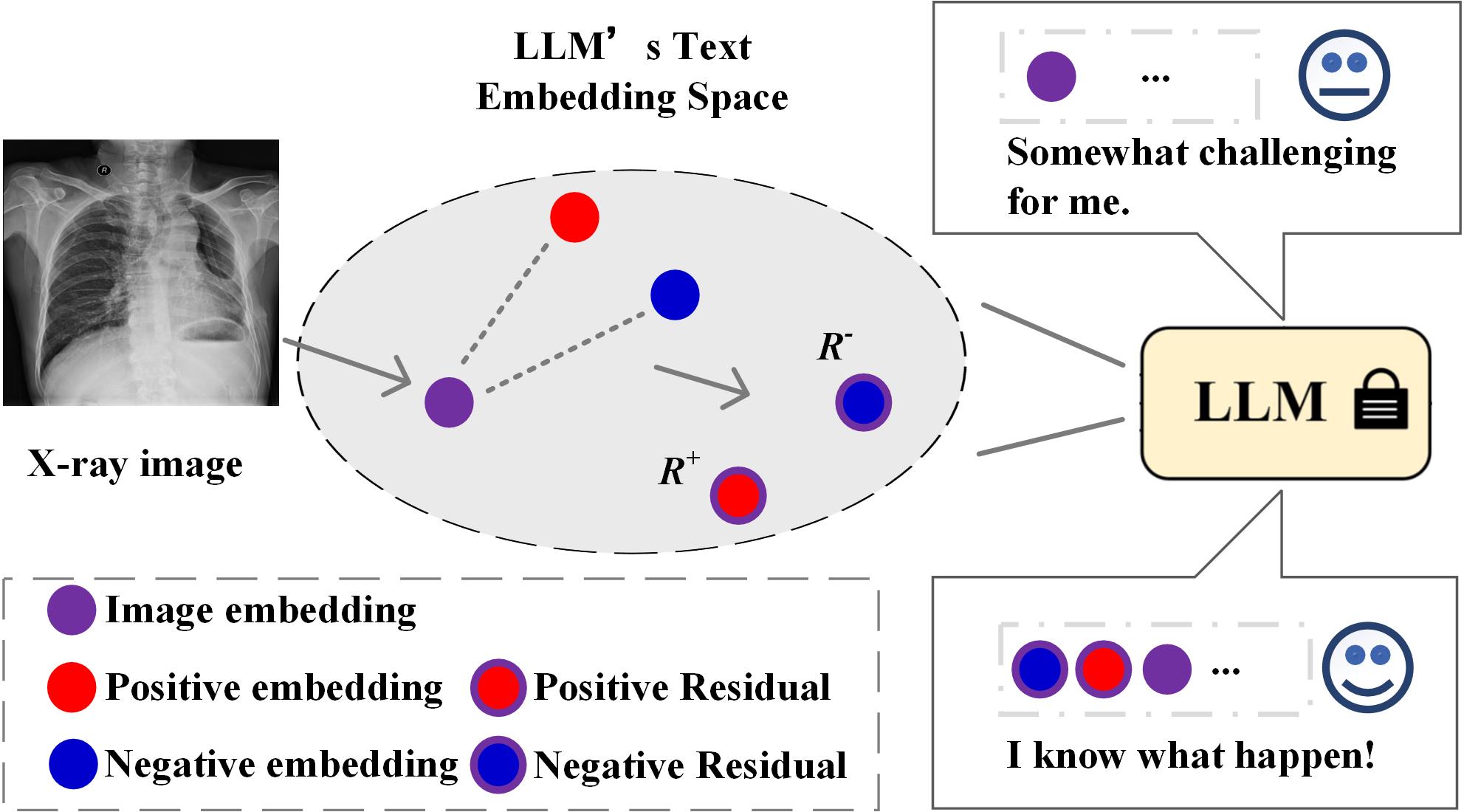}
    \caption{By concatenating the residual information with the current X-ray image as input prompt, the LLM leverages its intrinsic in-context learning capabilities to better understand the variations and generate more accurate medical reports.}
    \label{fig:residual_information}
\end{figure}

% \noindent $\bullet$ 
\subsubsection{Context Sample Retrieval}  

Given the visual tokens of the input X-ray images, one can directly feed them into the large language model to generate the medical reports, clearly, this approach will still reach its performance bottleneck. The context learning methods~\cite{wang2024BGFormer, dong2022ContextLearnsurvey} suggest that other samples in the training set may still play a positive role in each round of model optimization. Therefore, we consider designing a method to mine samples that are positively and negatively correlated with the current sample to enhance discriminative feature learning, thereby better guiding the LLM to generate accurate medical reports.

\noindent $\bullet$ \textbf{Positive and Negative Sample Selection.~} 
For each sample in a mini-batch, we can retrieve its context samples based on keywords in the medical reports. In our implementation, we exploit two different approaches:
1). We adopt the CheXbert~\cite{smit2020CheXbert} to find the possible 14 kinds of diseases from the annotated medical report. If the sample is annotated as \textit{No Finding}, we treat it as a negative context sample (without disease), otherwise positive (with disease). 
% 2). We observe that medical reports often use the word \textit{Note} to highlight significant findings or particularly crucial information, which typically indicates the presence of distinct diseases or key details in the corresponding images. 
% We believe that radiologists tend to emphasize important information in radiology images by including the word \textit{Note} in their reports, making such images suitable for clear differentiation in our sample selection process.
2). We observe that radiologists tend to emphasize significant findings in X-ray images by including the word \textit{Note} in their reports, making such images suitable for clear differentiation than other X-ray images in our sample selection process. To support our hypothesis, as illustrated in Fig.~\ref{fristIMG} (e), we used t-SNE to map the image features into a lower-dimensional space, and the results show a clear distinction between images associated with reports containing the word \textit{Note} and those without it. Accordingly, we classify context samples containing the word \textit{Note} as positive samples and those without it as negative samples, and then perform random sampling separately within each group.
% 2). We find that the medical report with and without \textit{Note} symbol can be roughly divided based on visual features, as illustrated in Fig.~\ref{fristIMG} (e). Similarly, we can treat the context samples with/without \textit{Note} as the positive/negative samples, respectively. 
% In addition, we also randomly select context samples to augment the training of our proposed R2GenCSR model. 

% After the context samples are retrieved, we extract the global visual features of X-ray images as $\{v_{g,1}, v_{g,2}, \dotsc, v_{g,n} \}, v_{g,i} \in \mathbb{R}^{E}$ using the Mamba backbone. These features are then projected into the language space of the LLM using a learnable projection layer, resulting in $\{c_{g,1}, c_{g,2}, \dotsc, c_{g,n}\}, c_{g,i} \in \mathbb{R}^{E}$. The projection layer are optimized during training to establish compatibility between the visual feature space and the LLM's language embedding space. 
After retrieving the context samples, we extract global visual features from the X-ray images using the same procedure as in Eq.~\ref{eq:v_g}. To distinguish these from the current training sample's feature $v_g$, we denote the context features as $\{c_{g,1}, c_{g,2}, \dotsc, c_{g,n}\}$, where each $c_{g,i} \in \mathbb{R}^{E}$ represents the $E$-dimensional embedding of the $i$-th context sample.

% Note that the Mamba vision backbone is frozen when extracting feature of context samples in each mini-batch. After that, the Mamba vision backbone is unfrozen. Freezing the network to extract context image features at the beginning of each epoch offers several benefits: 
% 1). It ensures stable feature extraction by maintaining consistent context image features and enhancing training efficiency and model performance. 
% 2). It reduces computational overhead by avoiding backpropagation during context feature extraction, saving training time and computational resources. This approach mirrors human cognitive processes of stable memory retrieval and difference analysis, making the model's behavior more human-like.

\noindent $\bullet$ \textbf{Residual Calculation.~} 
To guide the large language model to generate more accurate medical reports, we measure the difference between the current input X-ray sample and context samples, and term the difference as \textit{residual tokens}. For each context sample, we assign a disease prompt \textit{With disease} or \textit{Normal} and also take the visual-prompt difference into consideration. \modification{All quantities below are \emph{token embeddings} in the same projected language space $\mathbb{R}^E$. The residual tokens are calculated as follows:
\begin{align}
R_{v,i}^{+} &=  v_g - c_{g,i}^{+}, \quad
R_{v,i}^{-} = v_g - c_{g,i}^{-}, \\
\label{eq:r_negative}
R_{t}^+ &=  [v_g -t_{1}^+ ,v_g -t_{2}^+ ,\dotsc ,v_g -t_{p}^+],\\
R_{t}^- &=  [v_g -t_{1}^- ,v_g -t_{2}^- ,\dotsc ,v_g -t_{p}^-],
\label{eq:r_text}
\end{align}
In the above equations, $R_{v,i}^{+}$ and $R_{v,i}^{-}$ denote the residual tokens of the $i$-th positive and negative image, while $R_{t}^{+}$ and $R_{t}^{-}$ correspond to the residual tokens for the disease prompts. The notation [$\cdot , \cdot$] indicates the concatenation operation. 

To ensure the validity of directly subtracting textual prompts from visual embeddings in the LLM’s language space, we note that the prompts \textit{With disease} and \textit{Normal} are semantically opposed. Empirically, their average cosine similarity in the projection space after training is only 0.0165, indicating that the two prompts occupy well-separated regions. Thus, the subtraction operation can be interpreted as contrasting the observed image against alternative diagnostic categories.

To further couple textual and visual discrepancies, we insert $R_{v,i}$ into $R_{t}$ to form:
\begin{align}
R_{i}^{+} &= [v_g - t_1^+, v_g - t_2^+, \dots, R_{v,i}^{+}, \dots, v_g - t_p^+], \\
R_{i}^{-} &= [v_g - t_1^-, v_g - t_2^-, \dots, R_{v,i}^{-}, \dots, v_g - t_p^-],
\end{align}
where $R_{i}^{+}$ and $R_{i}^{-}$ denote the residual tokens corresponding to the $i$-th positive and negative context sample. %Conceptually, $R_{i}^{+}$ corresponds to a textual sequence such as: \textit{Note: $R_{v,i}^{+}$ normal}, while $R_{i}^{-}$ corresponds to: \textit{Note: $R_{v,i}^{-}$ with disease}.  

Therefore, the final prompt for the LLM is constructed by orderly concatenating the residuals and tokenized features as follows:
\begin{align}
\text{Prompt} & =  [R_{1}^{-} , R_{1}^{+} , R_{2}^{-} , R_{2}^{+} ,\dots, R_{n}^{+} , R_{n}^{+} , v_{s} , T], 
\label{eq:prompt_all}
\end{align}
Here, $T = \{t_1, t_2, \dots, t_k\}$ denotes the sequence of token embeddings obtained by processing the general instruction prompt with the LLM’s tokenizer. To leverage the LLM's in-context learning ability, we prioritize residuals by placing them at the beginning of the prompt sequence.}

As illustrated in Fig. \ref{fig:residual_information}, by capturing the semantic and visual discrepancies between the current image and the textual prompts, the residuals enable the LLM to better align visual and textual information. During the inference phase, the context sample pairs are selected from the training set, ensuring consistency with the training process.

% \noindent $\bullet$  
\subsubsection{LLM Head \& Loss Function} 
% The Large Language Model plays a central role in generating detailed and accurate medical reports from X-ray images. As described in the previous sections, the input to the LLM consists of a composite prompt that includes both visual and textual residuals derived from positive and negative context samples, along with the tokenized text corresponding to the input X-ray image. Once the contextual information has been integrated, the LLM generates the medical report more accuate. The generation process involves decoding the embedded and contextually enriched prompt into a coherent and comprehensive text.
The Large Language Model serves as the core component for generating comprehensive and precise medical reports from X-ray images. 
% By concatenating the residual information with the visual and textual representations of the current X-ray image, the LLM harnesses its inherent in-context learning capabilities to more effectively discern diagnostic variations and synthesize clinically accurate medical reports.
%%%% 
In our experiments, various LLMs are evaluated to achieve higher performance, including lightweight LLM Qwen1.5 (0.5B, 1.8B, 4B, 7B)~\cite{qwen}, the medium-sized Llama2 (7B)~\cite{touvron2023llama2}, Llama3 (8B)~\cite{dubey2024llama3}, and large-scale MedicalGPT (13B)~\cite{MedicalGPT}. For even larger LLMs, considering the computational cost, this paper will not consider them for the time being.

% \usepackage{multirow}
\begin{table*}[]
\caption{Comparison of our model’s performance on the \mmodification{IU X-Ray} and MIMIC-CXR datasets. The symbol $\dagger$ indicates that we follow the R2Gen annotation using \textit{Findings} and evaluate with our method because their report modifies the ground truth to an \textit{Impression} concatenated with \textit{Findings}. The best result is highlighted in bold, and the second-best result is underlined.}
\label{tab:results_iu_mimic}
\resizebox{\linewidth}{!}{
\begin{tabular}{c|l|c|ccccccc}
\hline 
\toprule [0.5 pt] 
\textbf{Dataset} & \textbf{Methods} & \textbf{Year} & \textbf{BLEU-1} & \textbf{BLEU-2} & \textbf{BLEU-3} & \textbf{BLEU-4} & \textbf{ROUGE-L} & \textbf{METEOR} & \textbf{CIDEr} \\ \hline
\multirow{13}{*}{\textbf{IU X-Ray}} 
 & R2Gen~\cite{chen2020R2Gen} & EMNLP 2020 & 0.470 & 0.304 & 0.219 & 0.165 & 0.371 & 0.187 & - \\
 & R2GenCMN~\cite{R2GenCMN} & ACL-IJCNLP 2021 & 0.475 & 0.309 & 0.222 & 0.170 & 0.375 & 0.191 & - \\
 & PPKED~\cite{liu2021exploring} & CVPR 2021 & 0.483 & 0.315 & 0.224 & 0.168 & 0.376 & 0.187 & 0.351 \\
 & AlignTrans~\cite{you2021aligntransformer} & MICCAI 2021 & 0.484 & 0.313 & 0.225 & 0.173 & 0.379 & 0.204 & - \\
 & Clinical-BERT~\cite{clinicalBert} & AAAI 2022 & 0.495 & 0.330 & 0.231 & 0.170 & 0.376 & 0.209 & 0.432 \\
 & METransformer~\cite{Wang_2023_CVPR} & CVPR 2023 & 0.483 & 0.322 & 0.228 & 0.172 & 0.380 & 0.192 & 0.435 \\
 & DCL~\cite{li2023dynamic} & CVPR 2023 & - & - & - & 0.163 & 0.383 & 0.193 & \textbf{0.586} \\
 % & Kiut~\cite{huang2023kiut} & CVPR 2023 & 0.525 & 0.360 & 0.251 & 0.185 & 0.409 & 0.242 & - \\
 % & COMG~\cite{COMG} & WACV 2024 & 0.536 & 0.378 & 0.275 & 0.206 & 0.383 & 0.218 & - \\
 % & MedM2G~\cite{zhan2024medm2g}* & CVPR 2024 & 0.533 & 0.369 & 0.278 & 0.212 & 0.416 & - & - \\
 % & EKAGen (RN-101)~\cite{bu2024instance} & CVPR 2024 & 0.526 & 0.361 & 0.267 & 0.203 & 0.404 & 0.214 & - \\
 & RAMT~\cite{RAMT} & IEEE TMM 2023 & 0.482 & 0.310 & 0.221 & 0.165 & 0.377& 0.195 & - \\ 
 & Token-Mixer~\cite{TokenMixer} & IEEE TMI 2024 & 0.483 & \underline{0.338} & \underline{0.250} & \underline{0.190} & \textbf{0.402} & 0.208 & 0.482 \\
 & PromptMRG~\cite{jin2024promptmrg} & AAAI 2024 & 0.401 & - & - & 0.098 & 0.160 & \textbf{0.281} & - \\
 & BootstrappingLLM~\cite{liu2024bootstrapping} & AAAI 2024 & \underline{0.499} & 0.323 & 0.238 & 0.184 & 0.390 & 0.208 & - \\ 
 % & MPMA~\cite{} & TMM 2024 & 0.518 & 0.337 & 0.253 & 0.179 & 0.388& 0.220 & - \\ 
 & FMVP~\cite{FVMP} & IEEE TMM 2024 & 0.485 & 0.315 & 0.225 & 0.169 & 0.398 & 0.201 & - \\ 
 & AHP-Full~\cite{AHP2025} & IEEE JBHI 2025 & 0.502 & 0.338 & 0.250 & 0.196 & 0.388 & 0.212 & 0.670 \\ 
% & HReMRG-MR ~\cite{HReMRG} & IEEE TNNLS 2025 & 0.440 & 0.306 & 0.214 & 0.149 & 0.381 & 0.197 & 0.524 \\
 \cline{2-10} 
 % & R2GenGPT~\cite{R2GenGPT} & Meta Radiology 2023 & 0.488 & 0.316 & 0.228 & 0.173 & 0.377 & 0.211 & 0.438 \\
 & R2GenGPT\textsuperscript{$\dagger$}~\cite{R2GenGPT} & Meta Radiology 2023 & 0.465 & 0.299 & 0.214 & 0.161 & 0.376 & \underline{0.219} & 0.542 \\
 & R2GenCSR & Ours & \textbf{0.514} & \textbf{0.351} & \textbf{0.262} & \textbf{0.206} & \underline{0.401} & 0.215 & \underline{0.579} \\
 % & R2GenCSR-Qwen & Ours & \textbf{0.514} & \textbf{0.351} & \textbf{0.262} & \textbf{0.206} & \underline{0.401} & 0.215 & \underline{0.579} \\
 % & R2GenCSR-Qwen & Ours & 0.516 & 0.354 & 0.265 & 0.210 & 0.401 & 0.217 & 0.593 \\ 
 \hline 
 % \toprule [0.5 pt] 
\multirow{13}{*}{\textbf{MIMIC-CXR}} 
 & R2Gen~\cite{chen2020R2Gen} & EMNLP 2020 & 0.353 & 0.218 & 0.145 & 0.103 & 0.277 & 0.142 & - \\
 & R2GenCMN~\cite{R2GenCMN} & ACL-IJCNLP 2021 & 0.353 & 0.218 & 0.148 & 0.106 & 0.278 & 0.142 & - \\
 & PPKED~\cite{liu2021exploring} & CVPR 2021 & 0.360 & 0.224 & 0.149 & 0.106 & 0.284 & 0.149 & 0.237 \\
 & AlignTrans~\cite{you2021aligntransformer} & MICCAI 2021 & 0.378 & 0.235 & 0.156 & 0.112 & 0.283 & 0.158 & - \\
 & Clinical-BERT~\cite{clinicalBert} & AAAI 2022 & 0.383 & 0.230 & 0.151 & 0.106 & 0.275 & 0.144 & 0.151 \\
 & METransformer~\cite{Wang_2023_CVPR} & CVPR 2023 & 0.386 & 0.250 & 0.169 & 0.124 & 0.291 & 0.152 & \textbf{0.362} \\
 & DCL~\cite{li2023dynamic} & CVPR 2023 & - & - & - & 0.109 & 0.284 & 0.150 & \underline{0.281} \\
 % & Kiut~\cite{huang2023kiut} & CVPR 2023 & 0.393 & 0.243 & 0.159 & 0.113 & 0.285 & 0.160 & - \\
 % & COMG~\cite{COMG} & WACV 2024 & 0.363 & 0.235 & 0.167 & 0.124 & 0.290 & 0.128 & - \\
 % & MedM2G~\cite{zhan2024medm2g}* & CVPR 2024 & 0.412 & 0.260 & 0.179 & 0.142 & 0.309 & - & - \\
 % & EKAGen (RN-101)~\cite{bu2024instance} & CVPR 2024 & 0.419 & 0.258 & 0.170 & 0.119 & 0.287 & 0.157 & - \\
 & RAMT~\cite{RAMT} & IEEE TMM 2023 & 0.362 & 0.229 & 0.157 & 0.113 & 0.284& 0.153 & - \\ 
 & Token-Mixer~\cite{TokenMixer} &IEEE TMI 2024 & \underline{0.409} & 0.257 & 0.175 & 0.124 & 0.288 & 0.158 & 0.163 \\
 & PromptMRG~\cite{jin2024promptmrg} & AAAI 2024 & 0.398 & - & - & 0.112 & 0.268 & 0.157 & - \\
 & BootstrappingLLM~\cite{liu2024bootstrapping} & AAAI 2024 & 0.402 & \underline{0.262} & \underline{0.180} & \underline{0.128} & \underline{0.291} & \textbf{0.175} & - \\ 
 % & ~\cite{} & AAAI 2024 & 0.402 & 0.xx & 0.xxx & 0.xxx & 0.xxx& 0.xxx & - \\ 
 % & ~\cite{} & AAAI 2024 & 0.402 & 0.xx & 0.xxx & 0.xxx & 0.xxx& 0.xxx & - \\ 
 & FMVP ~\cite{FVMP} & IEEE TMM 2024 & 0.389 & 0.236 & 0.156 & 0.108 & 0.284 & 0.150 & - \\ 
 & AHP-Full~\cite{AHP2025} & IEEE JBHI 2025 & 0.400 & 0.250 & 0.172 & 0.126 & 0.285 & 0.154 & 0.169 \\ 
 % & HReMRG-MR ~\cite{HReMRG} & IEEE TNNLS 2025 & 0.481 & 0.343 & 0.256 & 0.192 & 0.380 & 0.207 & 0.372 \\ \cline{2-10} 
 % & R2GenGPT~\cite{R2GenGPT} & Meta Radiology 2023 & 0.411 & 0.267 & 0.186 & 0.134 & 0.297 & 0.160 & 0.269 \\
  \cline{2-10} 
 & R2GenGPT\textsuperscript{$\dagger$}~\cite{R2GenGPT} & Meta Radiology 2023 & 0.408 & 0.256 & 0.174 & 0.125 & 0.285 & 0.167 & 0.244 \\
 & R2GenCSR & Ours & \textbf{0.420} & \textbf{0.268} & \textbf{0.186} & \textbf{0.136} & \textbf{0.291} & \underline{0.167} & 0.267 \\ 
 % & R2GenCSR-Llama2 & Ours & \textbf{0.420} & \textbf{0.268} & \textbf{0.186} & \textbf{0.136} & \textbf{0.291} & \underline{0.167} & 0.267 \\ 
 \hline 
 \toprule [0.5 pt] 
\end{tabular}
}
\end{table*}

To optimize our R2GenCSR framework, we adopt the cross-entropy loss function to measure the difference between the generated medical reports and the ground truth annotations. Specifically, we apply instruction-tuning to the LLM to generate medical reports, maintaining its original auto-regressive training objective. During this process, the LLM is fine-tuned specifically on the tokens of the medical report, guided by the instruction prompt that encapsulates the visual and textual residuals. Our loss function is defined as the negative log-likelihood of the sequence of report tokens. This can be formulated as:\modification{
\begin{equation}
    \label{lossFunction} 
    \mathcal{L} = -\sum _{i=1}^{L}\log p_{\theta } (y_{i} |Prompt ,Y_{r,< i} ).
\end{equation}
where $\theta $ denotes the trainable parameters, and $L$} is the length of the whole medical report. $y_{i}$ is the token being predicted at the current step \textit{i}, $Prompt$ is the instruction prompt that includes the residuals and tokenized features, and $Y_{r,< i}$ is the sequence of report tokens before the current prediction token $y_{i}$. The instruction-tuning ensures that the LLM generates a report that aligns with the provided instructions and context, thus producing a coherent and informative medical report.

\begin{table}[]
\caption{Comparison on the CheXpert Plus dataset.}
\label{tab:resultsCheXpertPlus}
\resizebox{\linewidth}{!}{
\begin{tabular}{l|ccccccc}
\hline \toprule [0.5 pt] 
\textbf{Method} & \textbf{Bleu-4} & \textbf{ROUGE-L} & \textbf{METEOR} & \textbf{CIDEr} \\ \hline
R2Gen~\cite{chen2020R2Gen}  & 0.091 & 0.262 & 0.131 & - \\
R2GenRL~\cite{qin2022R2GenRL} & 0.035& 0.186& 0.101& 0.012 \\
R2GenCMN~\cite{R2GenCMN}  & 0.095 & 0.262 & 0.142 & - \\
R2Gen-GPT~\cite{R2GenGPT}  & 0.102 & 0.267 & 0.157 & 0.179 \\
\hline 
% \textbf{R2GenCSR-Qwen1.5}  & 0.100 & 0.272 & 0.137 & 0.159 \\
R2GenCSR & \textbf{0.103} & \textbf{0.272} & \textbf{0.163} & \textbf{0.193} \\
% \textbf{R2GenCSR-Llama2} & \textbf{0.103} & \textbf{0.272} & \textbf{0.163} & \textbf{0.193} \\
\hline \toprule [0.5 pt] 
\end{tabular}
}
\end{table}

% \subsection{Clinical Efficacy Metrics}   

\begin{table}
\centering
\caption{Comparing the Clinical Efficacy (CE) metrics of different models on the Mimic-CXR dataset. }  
\label{tab:Clinical_Efficacy_Metrics}
\small 
\resizebox{\linewidth}{!}{ 
\begin{tabular}{l|c|cccc} 
\hline \toprule [0.5 pt]  
\textbf{Method} &\textbf{Year} &\textbf{Precision}  &\textbf{Recall} &\textbf{F1}  \\
\hline 

 R2Gen~\cite{chen2020R2Gen} & EMNLP 2020  & 0.333 & 0.273 & 0.276 \\
 METransformer~\cite{Wang_2023_CVPR} & CVPR 2023  & 0.364 & 0.309 & 0.311 \\
 KiUT~\cite{huang2023kiut} & CVPR 2023  & 0.371 & 0.318 & 0.321 \\
 % RGRG~\cite{Tanida2023RGRG} & CVPR 2023  & \{0} & \u} & \ \\
 R2GenGPT~\cite{R2GenGPT} & Meta Radiology 2023  & 0.392 & \underline{0.387} & \underline{0.389} \\
 RAMT~\cite{RAMT} & IEEE TMM 2023 & 0.380 & 0.342 & 0.335 \\ 
 DCL~\cite{li2023dynamic} & CVPR 2023 &\underline{0.471} &0.352 &0.373 \\
 MedRAT~\cite{hirsch2025medrat} & ECCV 2024  & 0.285 & 0.265 & 0.227 \\
 % CXR-IRGen~\cite{Shentu2024CXR-IRGen} & WACV 2024  & - & - & 0.293 \\
 HERGen~\cite{wang2024HERGen} & ECCV 2024  & 0.415 & 0.301 & 0.317 \\
 \hline
 R2GenCSR  &  Ours & \textbf{0.551}  & \textbf{0.432}  & \textbf{0.484}    \\
\hline \toprule [0.5 pt]  
\end{tabular}
} 
\label{tab:Clinical_Efficacy_Metrics}
\end{table}

% (a) False report of a finding in the candidate                     1.577
% (b) Missing a finding present in the reference                     2.256
% (c) Misidentification of a finding's anatomic location/position    0.156
% (d) Misassessment of the severity of a finding                     0.311
% (e) Mentioning a comparison that isn't in the reference            0.144
% (f) Omitting a comparison detailing a change from a prior study    0.098
% Name: mean, dtype: float64

% Clinical Efficacy (CE) metrics are widely used in the generation of medical reports to evaluate the relevance of medical terms. As illustrated in \tabref~\ref{tab:Clinical_Efficacy_Metrics}, we follow R2Gen~\cite{chen2020R2Gen}, DCL~\cite{li2023dynamic} to compute the micro-average scores over the 14 CheXpert observations for the Mimic-CXR dataset.
% Our model surpasses all existing methods in terms of Recall and F1 score, and achieves commendable performance in Precision, only slightly trailing behind HERGen~\cite{wang2024HERGen}.  Overall, our model demonstrates strong performance in CE metrics, reflecting its robustness and efficiency.
% /wangx/mimic-cxr-labeler/txt/validation/mimic_epoch8/pth_test36_mimic_cxr_validation_reports_chexpert_labeled.csv
% {'F1_MICRO': 0.4845034045550599,
%  'PRECISION_MICRO': 0.5515167713483897,
%  'RECALL_MICRO': 0.4320108866324715}

\section{Experiment}

\subsection{Datasets and Evaluation Metrics} 
We evaluate the performance of our model on three datasets, including \textbf{\mmodification{IU X-Ray}}~\cite{demner2016iuxray}, \textbf{MIMIC-CXR}~\cite{johnson2019mimicCXR}, and \textbf{CheXpert Plus}~\cite{chambon2024chexpertPLUS} dataset. A brief introduction to these datasets is given below. \modification{For all datasets, we use only the "Findings" section as the ground truth for evaluation.}

\noindent $\bullet$ \textbf{\mmodification{IU X-Ray}}~\cite{demner2016iuxray} is one of the most widely used publicly accessible medical image datasets in the field of medical report generation, which was released on year 2016. It contains 7,470 images and 3,955 radiology reports, each consisting of four parts: indication, comparison, \textit{Findings}, and \textit{Impression}. For a fair comparison, we have used the same dataset partition protocols as R2GenGPT and set the training/val/test for the dataset to 7:1:2.

\noindent $\bullet$ \textbf{MIMIC-CXR}~\cite{johnson2019mimicCXR} is a large publicly available dataset of chest radiographs with free-text radiology reports. These records, comprising 377, 110 radiographic images and 227, 835 radiology reports collected from 65, 379 individuals, span the years 2011-2016 and originate from the Beth Israel Deaconess Medical Center Emergency Department in Boston, MA. For a fair comparison, we used the same dataset partition protocols as R2GenGPT, where 270,790 samples were used to train the model, and another 2,130 and 3,858 samples were used as validation and test sets, respectively.

\noindent $\bullet$ \textbf{CheXpert Plus}~\cite{chambon2024chexpertPLUS} is a large, newly released, organized med dataset with 223K radiology report-X-ray pairs from 64.7K patients, with each report detailed into 11 sections and X-rays in DICOM format with 47 metadata elements. It's annotated for 14 chest conditions and patient metadata. We utilize the \textit{Findings} as our ground truth and randomly partition the dataset into a ratio of 7:1:2, which consists of training, validation, and testing sets with 40,463, 5,780, and 11,562 samples respectively. The split protocols for this dataset will be released for other researchers to reproduce our experiments.

\newcommand{\yes}{\ding{51}}   
\newcommand{\no}{\ding{55}} 
\begin{table*}[!htp]
\caption{ \mmodification{Component analysis of the key modules in our framework on MIMIC-CXR and IU X-Ray dataset. \textbf{B-1}, \textbf{B-4}, \textbf{R-L}, and \textbf{M} is short for BlEU-1, BlEU-4, ROUGE-L, METEOR, respectively. \textbf{P}, \textbf{R}, and \textbf{F1} is short for Precision, Recall, F1 score, respectively. }} 
\label{tab:component_mimic_iu}
\resizebox{\linewidth}{!}{
\begin{tabular}{c|c@{\ \ }c@{\ \ }c@{\ \ }c@{\ \ }c|cccccccc|cccc}
\hline
\toprule [0.5 pt]  
\multirow{3}{*}{\textbf{Index}} & \multirow{3}{*}{\textbf{VMamba}} & \multirow{3}{*}{\textbf{Context}} & \multirow{3}{*}{\textbf{Fixed pair}} & \multirow{3}{*}{\textbf{Qwen1.5}} & \multirow{3}{*}{\textbf{Llama2}} & \multicolumn{8}{c|}{\textbf{MIMIC-CXR}} & \multicolumn{4}{c}{\textbf{IU X-Ray}} \\ \cline{7-18} 
 &  &  &  &  &  & \multicolumn{4}{c|}{\textbf{NLG Metrics}} & \multicolumn{3}{c|}{\textbf{Clinical Efficacy}} & \multirow{2}{*}{\textbf{GREEN}} & \multicolumn{4}{c}{\textbf{NLG Metrics}} \\ \cline{7-13} \cline{15-18} 
 &  &  &  &  &  & \textbf{B-1} & \textbf{B-4} & \textbf{R-L} & \multicolumn{1}{c|}{\textbf{M}} & \textbf{P} & \textbf{R} & \multicolumn{1}{c|}{\textbf{F1}} &  & \textbf{B-1} & \textbf{B-4} & \textbf{R-L} & \textbf{M} \\ \hline
\#\textbf{1} & \no & \no & \no & \no & \yes & 0.407 & 0.126 & 0.282 & \multicolumn{1}{c|}{0.162} & 0.519 & 0.400 & \multicolumn{1}{c|}{0.452} & 0.316 & 0.454 & 0.173 & 0.376 & 0.210 \\
\#\textbf{2} & \yes & \no & \no & \no & \yes & 0.415 & 0.127 & 0.285 & \multicolumn{1}{c|}{0.165} & 0.552 & 0.430 & \multicolumn{1}{c|}{0.483} & 0.325 & 0.476 & 0.175 & 0.377 & 0.209 \\
\#\textbf{3} & \yes & \yes & \no & \no & \yes & 0.416 & 0.131 & 0.289 & \multicolumn{1}{c|}{0.167} & 0.544 & 0.423 & \multicolumn{1}{c|}{0.476} & 0.327 & 0.496 & 0.176 & 0.387 & 0.209 \\
\#\textbf{4} & \yes & \yes & \yes & \no & \yes & \textbf{0.420} & \textbf{0.136} & \textbf{0.291} & \multicolumn{1}{c|}{\textbf{0.167}} & \textbf{0.551} & \textbf{0.432} & \multicolumn{1}{c|}{\textbf{0.484}} & \textbf{0.329} & 0.483 & 0.180 & 0.385 & 0.219 \\
\#\textbf{5} & \no & \no & \no & \yes & \no & 0.327 & 0.110 & 0.282 & \multicolumn{1}{c|}{0.140} & 0.536 & 0.354 & \multicolumn{1}{c|}{0.426} & 0.298 & 0.478 & 0.180 & 0.384 & 0.195 \\
\#\textbf{6} & \yes & \no & \no & \yes & \no & 0.348 & 0.114 & 0.282 & \multicolumn{1}{c|}{0.144} & 0.526 & 0.361 & \multicolumn{1}{c|}{0.428} & 0.297 & 0.498 & 0.189 & 0.387 & 0.203 \\
\#\textbf{7} & \yes & \yes & \no & \yes & \no & 0.312 & 0.105 & 0.283 & \multicolumn{1}{c|}{0.138} & 0.551 & 0.366 & \multicolumn{1}{c|}{0.440} & 0.296 & 0.503 & 0.195 & 0.389 & 0.214 \\
\#\textbf{8} & \yes & \yes & \yes & \yes & \no & 0.359 & 0.116 & 0.283 & \multicolumn{1}{c|}{0.146} & 0.523 & 0.373 & \multicolumn{1}{c|}{0.436} & 0.305 & \textbf{0.514} & \textbf{0.206} & \textbf{0.401} & \textbf{0.215} \\ 
\hline  
\toprule [0.5 pt]
\end{tabular}
}
\end{table*}

For radiology report generation, we evaluate our model using four widely adopted natural language generation (NLG) metrics: CIDEr~\cite{vedantam_cider_2015}, BLEU~\cite{papineni_bleu_2002}, ROUGE-L~\cite{lin_rouge_2004}, and METEOR~\cite{banerjee_meteor_2005}. Specifically, CIDEr measures the consensus between generated and reference reports by computing the cosine similarity of n-grams, emphasizing the semantic relevance of medical descriptions. BLEU evaluates the precision of n-grams (up to 4-grams) in generated reports compared to reference reports, ensuring the accuracy of key medical terms and phrases. ROUGE-L focuses on the longest common subsequences between generated and reference reports, assessing the recall of critical clinical information. METEOR incorporates precision, recall, stemming, and synonymy matching, providing a balanced evaluation of the alignment between generated and reference reports. 

In addition to NLG metrics, we also evaluate clinical relevance in generated reports using Clinical Efficacy (CE)~\cite{liu2019CE_metric}. Following R2Gen~\cite{chen2020R2Gen} and DCL~\cite{li2023dynamic}, we extract disease labels from reports using the CheXpert labeler and compute micro-averaged Precision, Recall, and F1 scores on the MIMIC-CXR dataset.
% More details about these metrics can be found in the supplementary material. 

\begin{table*}[h!]
\caption{\mmodification{ \textbf{GREEN Metrics} of R2GenCSR. Clinically Significant Errors denote (a) False report of a finding in the candidate, (b) Missing a finding present in the reference, (c) Misidentification of a finding's anatomic location/position, (d) Misassessment of the severity of a finding, (e) Mentioning a comparison that isn't in the reference, (f) Omitting a comparison detailing a change from a prior study. Metrics with ↑ indicating higher is better and ↓ indicating lower is better.}}
\label{tab:green}
\resizebox{\linewidth}{!}{
\begin{tabular}{clccccccccc}
\hline 
\toprule [0.5 pt] 
% \multirow{2}{*}{\textbf{Dataset}} & 
% \multirow{2}{*}{\textbf{Method}} &  
% \multicolumn{6}{c}{\textbf{\#Clinically Significant Errors \downarrow}} & 
% \multirow{2}{*}{\textbf{\sum_{j=(a)}^{(f)} \#Error_j} \downarrow} &
% \multirow{2}{*}{\textbf{\#Matched Findings \uparrow}} &
% \multirow{2}{*}{\textbf{GREEN \uparrow}} \\ 

\multirow{2}{*}{\textbf{Dataset}} & 
\multirow{2}{*}{\textbf{Method}} &  
\multicolumn{6}{c}{\textbf{Clinically Significant Errors $\downarrow$}} &
\multirow{2}{*}{$\sum_{i=(a)}^{(f)} \#\,\textbf{error}_{\text{sig.}, i}$ $\downarrow$} &
\multirow{2}{*}{\textbf{\#\,matched findings $\uparrow$}} &
\multirow{2}{*}{\textbf{GREEN $\uparrow$}} \\ 
\cline{3-8} 
 & & (a) & (b) & (c) &  (d)  & (e)  & (f) &   &   &  \\ \hline
% \multirow{5}{*}{\textbf{\mmodification{IU X-Ray}}} 
% & R2Gen~\cite{chen2020R2Gen} & 0.581 & 1.327 & 0.036 & 0.061 & 0.086 & 0.025 & 2.1169 & 2.707 & 0.614\\
% & R2GenCMN~\cite{R2GenCMN} & - & 2.963 & 0.127 & 0.228 & \textbf{0.081} & 0.161 & 4.942 & 1.935 & 0.681   \\
% & CvT2DistilGPT2~\cite{CvT2DistilGPT2} & \textbf{1.150} & 2.881 & 0.146 & 0.234 & 0.103 & 0.153 & 4.666 & 1.967 & 0.313  \\
% & R2GenGPT~\cite{R2GenGPT} &- & \textbf{2.206} & 0.154 & 0.283 & 0.149 & \textbf{0.085} & \textbf{4.513} & 2.059 & 0.326\\
% \cline{2-11}
% & R2GenCSR & 0.442 & 1.269 & 0.022 & 0.036 & 0.110 & 0.020 & 1.900 & 2.834  &  0.653   \\
% \hline 
% \toprule [0.5 pt] 

\multirow{5}{*}{\textbf{MIMIC-CXR}} 
& R2Gen~\cite{chen2020R2Gen} & 1.310 & 3.089 & \textbf{0.103} & \textbf{0.201} & 0.082 & 0.142 & 4.926 & 1.803 & 0.283 \\
& R2GenCMN~\cite{R2GenCMN} & 1.383 & 2.963 & 0.127 & 0.228 & \textbf{0.081} & 0.161 & 4.942 & 1.935 & 0.297   \\
& CvT2DistilGPT2~\cite{CvT2DistilGPT2} & \textbf{1.150} & 2.881 & 0.146 & 0.234 & 0.103 & 0.153 & 4.666 & 1.967 & 0.313  \\
& R2GenGPT~\cite{R2GenGPT} &1.636 & \textbf{2.206} & 0.154 & 0.283 & 0.149 & \textbf{0.085} & \textbf{4.513} & 2.059 & 0.326\\
\cline{2-11}
& R2GenCSR & 1.577 & 2.256 & 0.156 & 0.311 & 0.144 & 0.098 & 4.541  & \textbf{2.099}  &  \textbf{0.329}    \\
\hline 
\toprule [0.5 pt] 

\multirow{5}{*}{\textbf{CheXpert Plus}} 
& R2Gen~\cite{chen2020R2Gen} & \textbf{1.235} & 3.638 & \textbf{0.150} & \textbf{0.122} & \textbf{0.021} & 0.103 & 5.269 & 0.938 & 0.168 \\
& R2GenCMN~\cite{R2GenCMN} & 1.512 & 3.336 & 0.218 & 0.165 & 0.043 & 0.126 & 5.399 & 1.179 & 0.197  \\
& CvT2DistilGPT2~\cite{CvT2DistilGPT2} & 1.815 & 3.289 & 0.252 & 0.182 & 0.091 & 0.132 & 5.760 & 1.166 & 0.185  \\
& R2GenGPT~\cite{R2GenGPT} & 1.928 & 2.361 & 0.280 & 0.262 & 0.108 & \textbf{0.059} & \textbf{4.998} & 1.521 & 0.244  \\
\cline{2-11}
& R2GenCSR & 2.015 & \textbf{2.311} & 0.258 & 0.251 & 0.121 & 0.069 & 5.024  & \textbf{1.535}   &  \textbf{0.246}    \\
\hline 
\toprule [0.5 pt] 
\end{tabular}
}
\end{table*}

\subsection{Implementation Details} 

% To assess the efficacy of our method, we employ the Llama2-7B and Qwen1.5-1.8B models as our large language models for the MIMIC-CXR and \mmodification{IU X-Ray} datasets, respectively. The Vmamba model served as the visual encoder for both datasets. 

In our experiments, the input X-ray image is default resized as $224 \times 224$, and the beam search is adopted for the report generation. The beam width is set as 5 and 3 for the \mmodification{IU X-Ray} and MIMIC-CXR datasets, respectively. The number of positive/negative sample pairs is set to 3. The training procedure is conducted on a server with NVIDIA A800 80GB GPUs using a mixed precision. We train the proposed R2GenCSR model for 20 and 25 epochs on the MIMIC-CXR and \mmodification{IU X-Ray} datasets, respectively. The mini-batch sizes are 36 and 32 for the MIMIC-CXR and \mmodification{IU X-Ray} datasets, and both models are trained using a learning rate of 1e-4. The CheXpert Plus dataset adopts the same training protocol as MIMIC-CXR. More details can be found in our source code.

\subsection{Comparison on Public Benchmarks}

\noindent $\bullet$ \textbf{Results on \mmodification{IU X-Ray} dataset. } 
As shown in \tabref~\ref{tab:results_iu_mimic}, we compare our results to state-of-the-art (SOTA) methods on the IU X-Ray datasets. It is important to note that the R2GenGPT method’s reported results were based on a concatenation of \textit{Impression} and \textit{Findings} as the testing ground truth, which we believe is not representative of general scenarios so we re-trained their method using only with \textit{Findings}. Our method demonstrates competitive performance, achieving a BLEU-4 score of 0.206, which surpasses existing methods, highlighting the effectiveness of our context-guided efficient radiology report generation framework. 
% Additionally, our approach attains in the precision of ROUGE-L, METEOR, and CIDEr scores, which are currently at 0.401, 0.412, and 0.579, respectively. These results, albeit not optimal, still affirm the superiority of our approach in generating precise and coherent medical reports compared to current SOTA methods.

\noindent $\bullet$ \textbf{Results on MIMIC-CXR dataset. } 
As shown in \tabref~\ref{tab:results_iu_mimic}, we report our results on the large-scale MIMIC-CXR dataset. Our method achieved a BLEU-1 score of 0.420, a BLEU-4 score of 0.136, and a ROUGE-L score of 0.291, indicating its ability to generate precise and contextually relevant medical reports. The CIDEr score of 0.267 suggests that the method produces descriptive reports that closely match the reference summaries, highlighting its practical application in clinical settings. Note that our approach employs a dataset-specific strategy, utilizing Llama2 as the text decoder, which is the model variant optimized for the MIMIC-CXR dataset, ensuring that the model is well-suited to the medical reports it processes. Notably, as illustrated in \tabref~\ref{tab:Clinical_Efficacy_Metrics}, our method achieves competitive CE scores for the MIMIC-CXR dataset, highlighting the robustness of our model in handling imbalanced datasets by effectively capturing relationships in both major and minor categories. 
% For more details on these metrics, please refer to the supplementary material.

\noindent $\bullet$ \textbf{Results on CheXpert Plus dataset.} 
As shown in \tabref~\ref{tab:resultsCheXpertPlus}, we re-train R2Gen series medical report generation models on the recently released CheXpert Plus dataset, including R2Gen~\cite{chen2020R2Gen}, R2GenRL~\cite{qin2022R2GenRL}, R2GenCMN~\cite{R2GenCMN}, and R2Gen-GPT~\cite{R2GenGPT}. The Llama2-based R2Gen-GPT outperforms others on all four metrics, but our proposed R2GenCSR model further surpasses R2Gen-GPT, achieving improvements of 0.001, 0.005, 0.006, and 0.014 on Bleu-4, ROUGE-L, METEOR, and CIDEr, respectively. These results validate the effectiveness of our proposed modules for radiology report generation.

\mmodification{\noindent $\bullet$ \textbf{Results on GREEN Metrics.~} 
Inspired by Liu et al.~\cite{liu2025MLRG} and GREEN~\cite{ostmeier-etal-2024-green}, GREEN evaluates factual correctness through explicit error-type matching, achieving close alignment with expert judgment. Since the GREEN metric was introduced recently and has not been widely adopted in earlier studies, 
we compare a limited set of methods.
We follow exactly the notation and summation format used in the original GREEN formulation. The GREEN score in the original formulation is defined as:
\begin{equation}
\text{GREEN} =
\frac{
\#\,\text{matched findings}
}{
\#\,\text{matched findings} +
\sum_{i=\text{(a)}}^{\text{(f)}} \#\,\text{error}_{\text{sig.}, i}
}
\label{equation:GREEN}
\end{equation}
\noindent Here, the subcategory labels (a)–(f) are the six predefined error types introduced in~\cite{ostmeier-etal-2024-green}, 
corresponding to
(a) false report of a finding;  
(b) missing a reference finding;  
(c) incorrect identification of anatomical location;  
(d) misassessment of severity;  
(e) spurious comparison not present in the reference;  
(f) omission of a comparison describing change from a prior study.

As shown in \tabref~\ref{tab:green}, R2GenCSR attains the highest GREEN scores across the evaluated datasets. This improvement is primarily driven by an increased number of matched findings, which improves the matched-to-error ratio in Eq.~\eqref{equation:GREEN}. While our model does not uniformly achieve the lowest count for every individual clinically significant error subcategory, its larger set of correctly identified findings yields a more favorable balance between correct findings and errors compared with competing methods. Overall, our model better preserves clinically relevant findings while maintaining comparable levels of clinically significant errors, leading to improved factual reliability of the generated reports. 
}

% \#Clinically Significant Errors represents the average number of clinically significant errors (lower is better); $\sum_{j=(a)}^{(f)}$ \#Error$_j$ denotes the average total errors (lower is better); \#Matched Findings indicates the average number of findings in the predicted report that match the ground truth (higher is better); and GREEN, computed using Eq.\eqref{equation:GREEN}, serves as the final evaluation metric (higher is better).
% \begin{equation}
% \text{GREEN} = \frac{\#\text{Matched Findings}_i}{\#\text{Matched Findings} + \sum_{i=(a)}^{(f)} \#\text{Error}_{i}}
% \label{equation:GREEN}
% \end{equation}
% As shown in \tabref~\ref{tab:green}, R2GenCSR attains the highest GREEN scores across the evaluated datasets. This improvement is primarily driven by an increased number of \#Matched Findings produced by R2GenCSR, which improves the matched-to-error ratio in Eq.~\eqref{equation:GREEN}. While our model does not uniformly achieve the lowest count for every individual clinically significant error subtype, its larger set of correctly identified findings yields a more favorable balance between correct findings and errors compared with competing methods. Overall, these results indicate that our model better preserves clinically relevant findings while maintaining comparable levels of clinically significant errors, leading to improved factual reliability of the generated reports.

\subsection{Component Analysis}
\mmodification{We conduct a detailed component analysis of our proposed framework on the IU X-Ray and MIMIC-CXR datasets to investigate the impact of each key module, including VMamba, the use of contextual information, the fixed-pair strategy, and the choice of language model, as shown in \tabref~\ref{tab:component_mimic_iu}. 
Specifically, replacing the Swin Transformer with VMamba yields richer visual representations and improved performance. The incorporation of contextual information facilitates the generation of higher-quality radiology reports, while the fixed-pair sampling strategy enhances metric scores by effectively leveraging both positive and negative samples.

Moreover, Qwen1.5 performs better than Llama2 on the IU X-Ray dataset. This can be explained by the scale-dependent behavior of large language models. Since IU X-Ray is relatively small, the smaller Qwen1.5 model is less sensitive to data sparsity and less prone to overfitting on limited patterns, allowing it to perform competitively or even better than larger models.
When all components are present, consistent improvement is observed across multiple evaluation metrics, which indicates that the method generalizes across different language models. Based on these comprehensive experiments, our method is effective and robust in generating radiology reports.

}

\begin{figure*}[!t]
    \centering
    \includegraphics[width=1\linewidth]{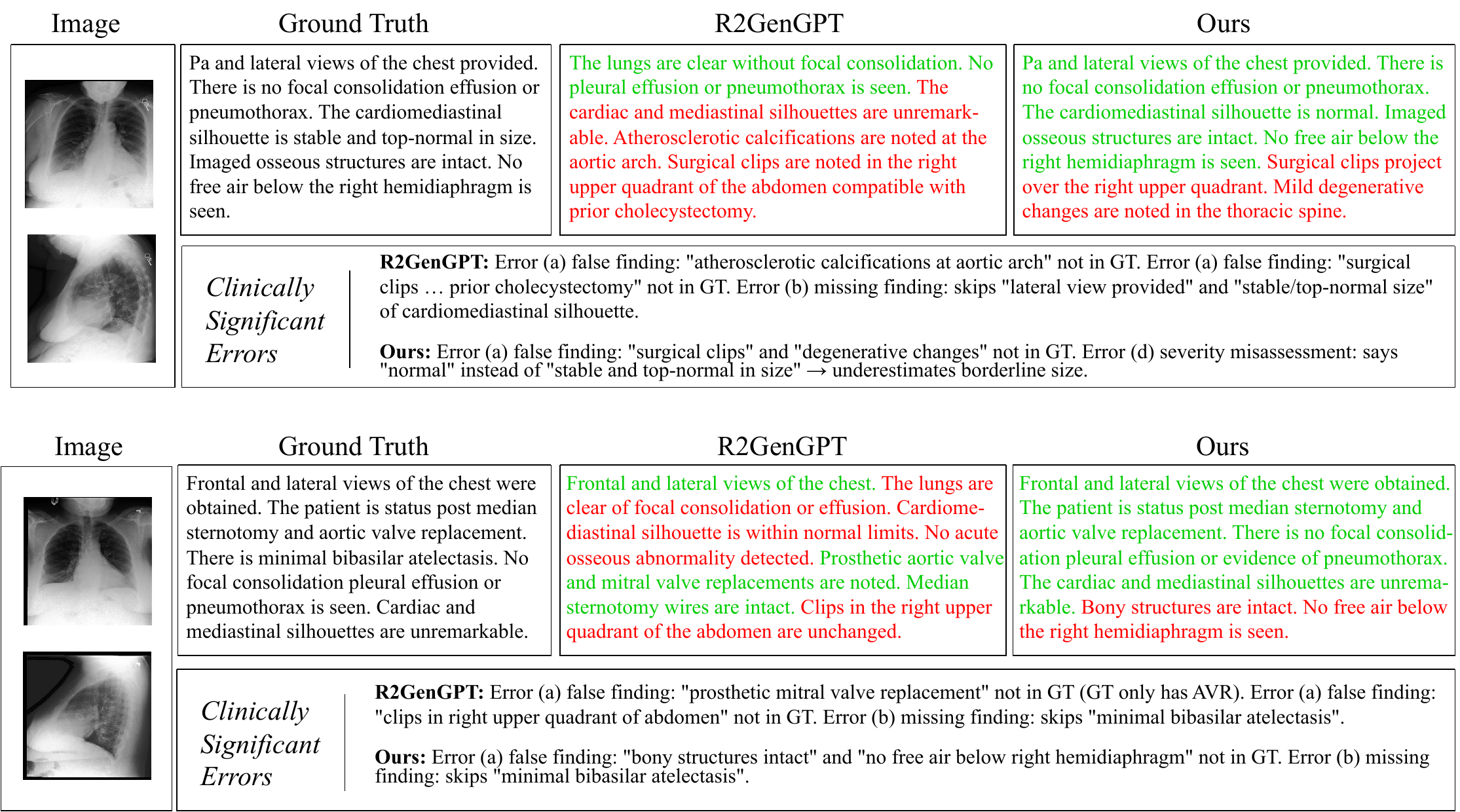}
    \caption{\mmodification{Comparison of X-ray images with corresponding ground truth reports and generated reports from R2GenGPT and our model on the MIMIC-CXR dataset. Sentences with semantically mismatched content are highlighted in red, while semantically matched sentences are highlighted in yellow. Clinically Significant Errors following the GREEN~\cite{ostmeier-etal-2024-green}: (a) False report of a finding in the candidate, (b) Missing a finding present in the reference, (c) Misidentification of a finding's anatomic location/position, (d) Misassessment of the severity of a finding, (e) Mentioning a comparison that isn't in the reference, (f) Omitting a comparison detailing a change from a prior study.}}%Comparison of X-ray images with corresponding ground truth reports and generated reports from both R2GenGPT (baseline) and our model on the MIMIC-CXR dataset. Sentences with semantically mismatched content are highlighted in red, while semantically matched sentences are highlighted in yellow.
    \label{fig:visual_report}
\end{figure*}

\subsection{Ablation Study}  
In this section, we conduct a series of ablation studies to evaluate the impact of various components within our proposed context-guided efficient radiology report generation framework. \mmodification{While IU X-Ray serves as the primary dataset for detailed component analysis due to its suitability for iterative validation, we additionally include key ablation experiments on the MIMIC-CXR dataset to confirm the consistency and generalizability of our findings.}

\begin{table}[!t]
% \caption{Vmamba on \mmodification{IU X-Ray} dataset. Beam search size is 5. Context samples pair is 3. Seed is 4096.}
\caption{Results of different VMamba model scales on the \mmodification{IU X-Ray} dataset.} 
\label{tab:abl_vmamba_scale_iu}
\resizebox{\linewidth}{!}{
\begin{tabular}{l|cccc}
\hline \toprule [0.5 pt]  
\textbf{Method} & \textbf{Bleu-1} & \textbf{Bleu-4} & \textbf{ROUGE-L} & \textbf{METEOR} \\ \hline
\textbf{Tiny} & 0.484 & 0.196 & 0.398 & 0.205 \\
\textbf{Small} &0.505&0.204&0.391&0.212\\
\textbf{Base} & 0.514 & 0.206 & 0.401 & 0.215 \\
\hline \toprule [0.5 pt]  
\end{tabular}
}
\end{table}

\begin{table}[!t]
\caption{\modification{Results of different context retrieve methods on \mmodification{IU X-Ray}.}} 
\label{tab:abl_retrieve_iu}
\resizebox{\linewidth}{!}{
\begin{tabular}{l|cccc}
\hline \toprule [0.5 pt]  
\textbf{Method} & \textbf{Bleu-1} & \textbf{Bleu-4} & \textbf{ROUGE-L} & \textbf{METEOR} \\ \hline
\textbf{Chexbert}& 0.495 & 0.195 & 0.391 & 0.207 \\
\textbf{Random} & 0.504 ± 0.004 & 0.197 ± 0.002  & 0.390 ± 0.002 & 0.213 ± 0.004 \\
\textbf{Keyword} & 0.514 & 0.206 & 0.401 & 0.215 \\
\hline \toprule [0.5 pt]  
\end{tabular}
}
\end{table}
% [A{'Bleu_1': 0.501, 'Bleu_2': 0.33600669814948037, 'Bleu_3': 0.24925890220747976, 'Bleu_4': 0.194, 'ROUGE_L': 0.388, 'METEOR': 0.209, }
% [A{'Bleu_1': 0.509, 'Bleu_2': 0.34259737721688244, 'Bleu_3': 0.25443773645928036, 'Bleu_4': 0.198, 'ROUGE_L': 0.391, 'METEOR': 0.214, }
% 0.502 & 0.198  & 0.392 & 0.217
% Bleu_1: 0.504 ± 0.004
% Bleu_4: 0.197 ± 0.002
% ROUGE_L: 0.390 ± 0.002
% METEOR: 0.213 ± 0.004

\begin{table}[!t]
\caption{Ablation of context samples on \mmodification{IU X-Ray} dataset. }
\label{tab:abl_context_fixnum_iu}
\resizebox{\linewidth}{!}{
\begin{tabular}{c|cccc}
\hline \toprule [0.5 pt]  
\textbf{\#Pairs} & \textbf{Bleu-1} & \textbf{Bleu-4} & \textbf{ROUGE-L} & \textbf{METEOR} \\ \hline
\textbf{1} & 0.513 & 0.196 & 0.390 & 0.215 \\
\textbf{3} & 0.514 & 0.206 & 0.401 & 0.215 \\
\textbf{5} & 0.490 & 0.182 & 0.375 & 0.203 \\
\textbf{10} & 0.491 & 0.182 & 0.383 & 0.205 \\
\hline \toprule [0.5 pt]  
\end{tabular}
}
\end{table}

\noindent $\bullet$ \textbf{Analysis of different VMamba backbones.~} 
As shown in \tabref~\ref{tab:abl_vmamba_scale_iu}, we evaluate the Tiny, Small, and Base versions of VMamba on the \mmodification{IU X-Ray} dataset. As the scale of the VMamba backbone increases, the performance metrics show a consistent improvement. The Base version of VMamba demonstrates the best overall performance, with improvements over the Small version of 0.009, 0.002, 0.010, and 0.003 in Bleu-1, Bleu-4, ROUGE-L, and METEOR, respectively. It confirms that a larger VMamba backbone can capture more intricate visual features from X-ray images, thereby enhancing the quality of the generated medical reports.

\noindent $\bullet$ \textbf{Analysis of different context sample retrieve methods.~}
As shown in \tabref~\ref{tab:abl_retrieve_iu}, we assess the performance of three distinct methods: Chexbert~\cite{smit2020CheXbert} is utilized to extract 14 medical observations, we define the images corresponding to reports labeled "\textit{No Finding}" serving as negative samples, while all other images are classified as positive. Despite its moderate performance, Chexbert provides a baseline for comparison. The Random method retrieves context samples in a completely random manner, without any division into positive or negative groups, and it leads to a slight improvement in the generation results. \modification{To ensure fairness and avoid accidental bias, we report its performance as mean ± standard deviation over multiple runs with six randomly selected samples.} The Keyword method retrieves samples based on the presence or absence of the word "\textit{Note}" in the reports, as analyzed in the "Positive and Negative Sample Selection" subsection. 
% The Keyword method retrieves samples based on keyword matching and is the most effective approach, surpassing the other two methods on all evaluated metrics.

\noindent $\bullet$ \textbf{Analysis of the different number of context samples.~} The \tabref~\ref{tab:abl_context_fixnum_iu} illustrates the performance across different numbers of context sample pairs. The results reveal an optimal point at 3 context sample pairs, which achieves higher scores than 10 pairs, with improvements for Bleu-1, Bleu-4, ROUGE-L, and METEOR of 0.023, 0.024, 0.018, and 0.010, respectively. As the number of context samples increases (from 3 to 10), there is a noticeable decline in performance across all metrics. We suggest that while additional context samples can contribute to a richer feature representation, an excessive number may introduce noise, detracting from the model’s ability to generate accurate and coherent reports. The use of a single context sample pair also yields competitive results, but the slight improvement seen with three pairs indicates that a moderate amount of context can enhance the learning process.

\noindent $\bullet$ \textbf{Analysis of different stages for feature subtraction.~} 
As illustrated in \tabref~\ref{tab:abl_residual_stage_iu}, conducting visual subtraction in the large language model embedding space yields better results than subtraction outside of it, and when both visual and context-instructed text residuals are considered, all evaluation metrics show improvements after the LLM projection space, highlighting the importance of feature subtraction at different stages and confirming the effectiveness of the proposed approach in enhancing feature representation.

\begin{table}[!t]
\caption{Feature subtraction at different stages on the \mmodification{IU X-Ray} dataset. ‘Before-proj. VR’ denotes the visual residual conducted before projection, ‘After-proj. VR’ denotes the visual residual conducted after projection, and ‘After-proj. VR + TR’ denotes the visual and context-instructed text residuals conducted after projection.}
\label{tab:abl_residual_stage_iu}
\resizebox{\linewidth}{!}{
\begin{tabular}{l|cccc}
\hline \toprule [0.5 pt]  
\textbf{Method} & \textbf{Bleu-1} & \textbf{Bleu-4} & \textbf{ROUGE-L} & \textbf{METEOR} \\ \hline
\textbf{Before-proj. VR} & 0.503 & 0.196& 0.390&0.211\\
\textbf{After-proj. VR} & 0.506 & 0.200 & 0.392 & 0.213 \\
\textbf{After-proj. VR + TR} & 0.514 & 0.206 & 0.401 & 0.215 \\  
\hline \toprule [0.5 pt]  
\end{tabular}
}
% \label{tab:my-table}
\end{table}

\begin{table}[t]
\caption{Results under different resolutions on \mmodification{IU X-Ray} dataset.}  
\label{tab:abl_image_scale_iu}
\resizebox{\linewidth}{!}{
\begin{tabular}{l|cccc}
\hline \toprule [0.5 pt]  
\textbf{Scale} & \textbf{Bleu-1} & \textbf{Bleu-4} & \textbf{ROUGE-L} & \textbf{METEOR} \\ \hline
\textbf{512 $\times$ 512} & 0.500 & 0.196 & 0.388 & 0.205 \\
\textbf{448 $\times$ 448} & 0.503 & 0.201 & 0.385 & 0.209 \\
\textbf{224 $\times$ 224} & 0.514 & 0.206 & 0.401 & 0.215 \\
\hline \toprule [0.5 pt]  
\end{tabular}
}
\end{table}

\begin{table}[h]
\caption{Evaluation between VMamba and Swin Transformer on MIMIC-CXR dataset.}
\label{tab:abl_vmamba_swin_compara_iu}
\resizebox{\linewidth}{!}{
\begin{tabular}{c|cc}
\hline \toprule [0.5 pt]  
\textbf{Scale and Efficiency} & \textbf{R2GenCSR-VMamba} & \textbf{R2GenCSR-SwinT} \\ \hline
Trainable Parameter & 91.7 M & 90.9 M \\ 
FLOPs & 1852.35 G & 1855.02 G \\
Training Time &  3.98 h/Epoch & 5.85 h/Epoch\\
Testing Time &  0.52 h& 0.52 h \\
Memory & 75723 MB & 70203 MB \\
\hline \toprule [0.5 pt]  
\end{tabular}
}
\end{table}

\noindent $\bullet$ \textbf{Analysis of different resolutions for report generation.~} 
As shown in \tabref~\ref{tab:abl_image_scale_iu}, we conducted experiments on the \mmodification{IU X-Ray} dataset using three image resolutions: 512 $\times$ 512, 448 $\times$ 448, and 224 $\times$ 224. The performance metrics, including Bleu-1, Bleu-4, ROUGE-L, and METEOR, improved by 0.014, 0.010, 0.013 and 0.010 as the resolution decreased from 512 $\times$ 512 to 224 $\times$ 224, respectively. This unexpected trend can be partially explained by the fact that we utilized the Vmamba model which was pre-trained on 224 $\times$ 224 resolution and it is adept at processing and extracting meaningful features from images of this specific resolution.
% However, it is crucial to note that while lower resolutions may offer computational benefits, they also pose a risk of omitting critical diagnostic details. 

\begin{table*}[!htp]
\caption{\mmodification{Ablation study using only positive or only negative samples on MIMIC-CXR and IU X-Ray. \textbf{Pos.} denotes positive samples, and \textbf{Neg.} denotes negative samples.}}
\label{tab:pos_neg}
\resizebox{\linewidth}{!}{
\begin{tabular}{cc|cccccccc|cccc}
\hline
\toprule [0.5 pt]
\multirow{3}{*}{\textbf{Pos.}} & \multirow{3}{*}{\textbf{Neg.}} & \multicolumn{8}{c|}{\textbf{MIMIC-CXR}} & \multicolumn{4}{c}{\textbf{IU X-Ray}} \\ \cline{3-14} 
 &  & \multicolumn{4}{c|}{\textbf{NLG Metrics}} & \multicolumn{3}{c|}{\textbf{Clinical Efficacy}} & \multirow{2}{*}{\textbf{GREEN}} & \multicolumn{4}{c}{\textbf{NLG Metrics}} \\ \cline{3-9} \cline{11-14} 
 &  & \textbf{B-1} & \textbf{B-4} & \textbf{R-L} & \multicolumn{1}{c|}{\textbf{M}} & \textbf{P} & \textbf{R} & \multicolumn{1}{c|}{\textbf{F1}} &  & \textbf{B-1} & \textbf{B-4} & \textbf{R-L} & \textbf{M} \\ \hline
\no & \no & 0.407 & 0.126 & 0.282 & \multicolumn{1}{c|}{0.162} & 0.519 & 0.400 & \multicolumn{1}{c|}{0.452} & 0.316 & 0.498 & 0.189 & 0.387 & 0.203 \\
\yes & \no & 0.417 & 0.133 & 0.288 & \multicolumn{1}{c|}{0.167} & 0.546 & 0.426 & \multicolumn{1}{c|}{0.479} & 0.328 & 0.485 & 0.195 & 0.380 & 0.203 \\
\no & \yes & 0.415 & 0.133 & 0.287 & \multicolumn{1}{c|}{0.167} & 0.539 & 0.435 & \multicolumn{1}{c|}{0.481} & 0.331 & 0.505 & 0.190 & 0.372 & 0.207 \\
\yes & \yes & 0.420 & 0.136 & 0.291 & \multicolumn{1}{c|}{0.167} & 0.551 & 0.432 & \multicolumn{1}{c|}{0.484} & 0.329 & 0.514 & 0.206 & 0.401 & 0.215 \\ \hline
\toprule [0.5 pt]
\end{tabular}
}
\end{table*} 

% \begin{table*}[!htp]
% \caption{\mmodification{Ablation study using only positive or only negative samples on MIMIC-CXR. Pos. denotes positive samples, and Neg. denotes negative samples.}}
% \label{tab:pos_neg_mimic}
% \resizebox{\linewidth}{!}{
% \begin{tabular}{c|c|cccc|ccc|c}
% \hline 
% % \toprule [0.5 pt]  
% \multirow{2}{*}{\textbf{Pos.}} & \multirow{2}{*}{\textbf{Neg.}} & \multicolumn{4}{c|}{\textbf{NLG Metrics}} & \multicolumn{3}{c|}{\textbf{Clinical Efficacy}} & \multirow{2}{*}{\textbf{GREEN}} \\ \cline{3-9}
%  &  &   \textbf{Bleu-1} & \textbf{Bleu-4} & \textbf{ROUGE-L} & \textbf{METEOR} & \textbf{Precision} & \textbf{Recall} & \textbf{F1} &  \\ \hline
% \no & \no & 0.407 & 0.126 & 0.282 & 0.162 & 0.519 & 0.400 & 0.452 & 0.316  \\
% \yes & \no & 0.417 & 0.133 & 0.288 & 0.167 & 0.546 & 0.426 & 0.479 & 0.328  \\
% \no & \yes & 0.415 & 0.133 & 0.287 & 0.167 & 0.539 & 0.435 & 0.481 & 0.331 \\
% \yes & \yes & 0.420 & 0.136 & 0.291 & 0.167 & 0.551 & 0.432 & 0.484 & 0.329  \\ \hline
% % \toprule [0.5 pt]
% \end{tabular}
% }
% \end{table*} 

\mmodification{
\noindent $\bullet$ \textbf{Analysis of Positive vs. Negative Samples.~}  
As shown in Table~\ref{tab:pos_neg}, using only positive samples or only negative samples already brings clear improvements over the baseline without residual guidance, confirming that each component individually contributes to enhancing report generation. Positive samples improve clinical efficacy metrics, while negative samples slightly improve GREEN score. Importantly, combining both positive and negative samples yields the best overall performance across all metrics, indicating their complementary effects in guiding the LLM to generate more accurate and comprehensive reports. 
}

\newcommand{\ImgWarpLeft}{\textless{}Img\textgreater{}}
\newcommand{\ImgWarpRight}{\textless{}/Img\textgreater}
\newcommand{\ImgWarp}[1]{\ImgWarpLeft#1\ImgWarpRight}
\newcommand{\InstructiontWrap}[1]{\texttt{{Human}}: \ImgWarp{$v_{s}$} #1 \textbackslash{n}\texttt{{Assistant}}:}
\noindent $\bullet$ \textbf{Analysis of Positive-Negative context-instructed text and Instruction prompt.~}
As shown in \tabref~\ref{tab:abl_instruction_prompt_iu}, we examine the influence of positive-negative context-instructed text and different instruction prompts on our model’s report generation quality. For instance, by modifying the context-instructed text from \textit{Note: \ImgWarp{$R_{v}^{-}$} normal. Note: \ImgWarp{$R_{v}^{+}$} with disease.} to \textit{Observation: \ImgWarp{$R_{v}^{-}$} appears to be normal and healthy. Observation: \ImgWarp{$R_{v}^{+}$} shows clear signs of disease.}, the Bleu-4 score decreased by 0.010. Similarly, by changing the instruction prompt from \textit{Generate a comprehensive and detailed diagnosis report for this chest x-ray image.} to \textit{Construct a full and methodical diagnostic summary for the chest X-ray displayed.}, the Bleu-4 score decreased by 0.007. \modification{These minor variations stem from the LLM’s sensitivity to prompt wording. Concise and medically conventional phrasing aligns better with its pretrained medical language distribution.} Although the differences in performance between the prompts are relatively small, optimizing prompts can subtly affect the model’s output. 

\begin{table*}[!ht]
\caption{Analysis of prompt design variants on \mmodification{IU X-Ray}. For interpretability, we replace text residuals
 $R_{t}^{-}$, $R_{t}^{+}$ and general instruction prompt $T$ with their original text descriptions.}
\label{tab:abl_instruction_prompt_iu}
\centering
% \begin{center}
\resizebox{\linewidth}{!}{
\begin{tabular}{l|cccc}
% \begin{tabular}{p{10cm}|cccc}
\hline \toprule [0.5 pt]  
\textbf{ \ Method} & \textbf{Bleu-1} & \textbf{Bleu-4} & \textbf{ROUGE-L} & \textbf{METEOR} \\ \hline
\begin{tabular}[c]{p{10cm}}Note: \ImgWarp{$R_{v}^{-}$} normal. Note: \ImgWarp{$R_{v}^{+}$} with disease. \InstructiontWrap{Generate a comprehensive and detailed diagnosis report for this chest xray image.}\end{tabular} &0.514&0.206&0.401&0.215 \\ \hline
\begin{tabular}[c]{p{10cm}}Observation: \ImgWarp{$R_{v}^{-}$} appears to be normal and healthy. Observation: \ImgWarp{$R_{v}^{+}$} shows clear signs of disease. \InstructiontWrap{Generate a comprehensive and detailed diagnosis report for this chest x-ray image.}\end{tabular}  & 0.503 & 0.196 & 0.384 & 0.209 \\ \hline
\begin{tabular}[c]{p{10cm}}Indication: \ImgWarp{$R_{v}^{-}$} shows no signs of pathology. Indication: \ImgWarp{$R_{v}^{+}$} exhibits symptoms of a medical condition. \InstructiontWrap{Generate a comprehensive and detailed diagnosis report for this chest x-ray image.}\end{tabular}  & 0.502 & 0.200 & 0.393 & 0.211 \\ \hline
\begin{tabular}[c]{p{10cm}}Findings: \ImgWarp{$R_{v}^{-}$} reveals a lack of abnormalities. Findings: \ImgWarp{$R_{v}^{+}$} reveals the presence of a pathology. \InstructiontWrap{Generate a comprehensive and detailed diagnosis report for this chest x-ray image.}\end{tabular}  & 0.499 & 0.202 & 0.393 & 0.207 \\ \hline
% Note: \ImgWarp{$R_{v}^{+}$} normal. Note: \ImgWarp{$R_{v}^{+}$} with disease. & 0.517 & 0.206 & 0.402 & 0.219 \\
\begin{tabular}[c]{p{10cm}}Note: \ImgWarp{$R_{v}^{-}$} normal. Note: \ImgWarp{$R_{v}^{+}$} with disease. \InstructiontWrap{Construct a full and methodical diagnostic summary for the chest X-ray displayed.}\end{tabular} & 0.497 & 0.199 & 0.392 & 0.209 \\ \hline
\begin{tabular}[c]{p{10cm}}Note: \ImgWarp{$R_{v}^{-}$} normal. Note: \ImgWarp{$R_{v}^{+}$} with disease. \InstructiontWrap{Develop a detailed and professional medical assessment from this chest X-ray image.}\end{tabular} & 0.504 & 0.196 & 0.393 & 0.217 \\ \hline 
\begin{tabular}[c]{p{10cm}}Note: \ImgWarp{$R_{v}^{-}$} normal. Note: \ImgWarp{$R_{v}^{+}$} with disease. \InstructiontWrap{Analyze and generate a detailed report on the findings of this chest X-ray.}\end{tabular} & \ 0.502 & 0.203 & 0.388 & 0.213 \\  
\hline \toprule [0.5 pt]  
\end{tabular}
}
% \end{center}
\end{table*}

\begin{figure}[!t]
    \centering
    \includegraphics[width=1\linewidth]{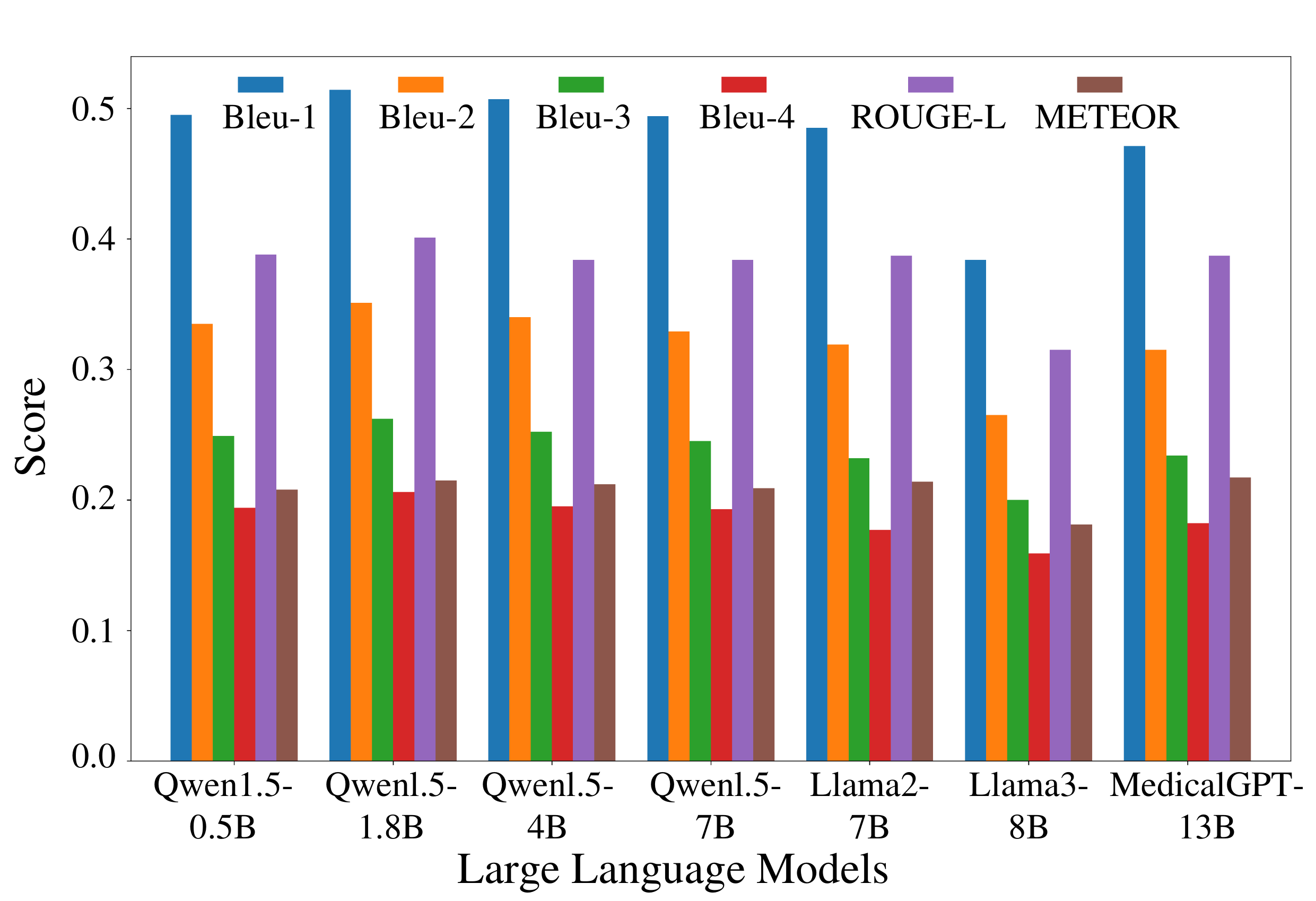}
    \caption{\mmodification{Performance comparison of LLMs on the IU X-Ray dataset, revealing significant variations across models with the Qwen series demonstrating superior performance on average.}}
    \label{fig:abl_model_type_iu}
\end{figure}

\noindent $\bullet$ \textbf{Analysis of different LLMs for report generation.~} 
As illustrated in Fig.~\ref{fig:abl_model_type_iu}, we present the performance among models of varying sizes and architectures, including the Qwen and Llama series models. The Qwen series models demonstrate superior performance on average, outperforming other models across multiple evaluation metrics. Specifically, the Qwen1.5-1.8B model demonstrates a notable improvement in metrics such as Bleu-4 and ROUGE-L by 0.206 and 0.401, respectively. However, the Llama3-8B model, despite its larger size, underperforms compared to the Qwen1.5-4B model, even worse than Llama2-7B on \mmodification{IU X-Ray} dataset. The specialized MedicalGPT (Llama-13B) model exhibits competitive performance within the Llama series, showcasing the potential benefits of domain-specific fine-tuning for large language models in medical report generation tasks. Therefore, selecting an appropriate LLM that balances size, architecture, and domain specialization is necessary for optimal report generation performance.

% \noindent $\bullet$ \textbf{Analysis of VMamba and Swin-Transformer for report generation.~}
% In \tabref~\ref{tab:abl_vmamba_swin_compara_iu}, we compare the training and testing efficiency of VMamba and Swin Transformer as vision backbones on a single A800 80G GPU. Although R2GenCSR-VMamba has slightly more trainable parameters (91.7 M vs. 90.9 M) and consumes more memory (75723 MB vs. 70203 MB), its training time requires only 3.98 hours per epoch, which is less than the 5.85 hours per epoch needed for R2GenCSR-Swin. Despite similar testing times for both models, VMamba has a lower FLOPs count (1852.35 G) than Swin Transformer (1855.02 G) and the overall efficiency of R2GenCSR-VMamba is superior.

\noindent $\bullet$ \textbf{Analysis of VMamba and Swin-Transformer for report generation.~}
\modification{
In \tabref~\ref{tab:abl_vmamba_swin_compara_iu}, we compare VMamba and Swin Transformer as vision backbones on a single A800 80G GPU. While R2GenCSR-VMamba has slightly more trainable parameters (91.7 M vs. 90.9 M) and higher memory usage (75,723 MB vs. 70,203 MB), it reduces training time per epoch from 5.85 hours to 3.98 hours. The FLOPs of VMamba are comparable to Swin Transformer (1852.35 G vs. 1855.02 G), and testing times are similar for both models. These results indicate that, despite the modest increase in memory usage, VMamba provides better training efficiency and accuracy trade-off for R2GenCSR.

% \mmmodification{The performance gain of VMamba over Swin Transformer stems not merely from faster training, but from its architectural capacity for efficient global modeling. Swin Transformer relies on local window attention and multi-layer shifting to propagate information across regions using an indirect and lossy approximation of global context. In contrast, Mamba’s selective state space model (Eq. \ref{eq:mamba}) maintains a recurrent hidden state $h_t$ that implicitly aggregates information from all previous tokens in O(N) time. Crucially, the input dependent parameters $\bar{B}(x_t)$ and $C(x_t)$ allow the model to dynamically focus on clinically relevant regions while preserving global anatomical structure. This is particularly beneficial in chest X-ray interpretation, where findings often involve spatially distant but semantically linked regions. Thus, Mamba’s linear complexity is not just an efficiency trick, it enables a more faithful and scalable representation of medical images.}
}

\modification{
\begin{table}[!htp]
\caption{\modification{Analysis of backbone and decoder variants. "Transf." denotes a 3-layer, 8-head Transformer used as the language decoder.}}
\label{tab:basc_transformer}
\resizebox{\linewidth}{!}{
\begin{tabular}{l|cccc}
\hline 
% \toprule [0.5 pt]  
 \textbf{Method} & \textbf{Bleu-1} &  \textbf{Bleu-4} & \textbf{ROUGE-L} & \textbf{METEOR}  \\ \hline
 R2Gen& 0.470 & 0.165& 0.371& 0.187\\
 R2Gen-Mamba& 0.482& 0.176& 0.382&0.208  \\
 Vmamba + Transf.& 0.493& 0.176& 0.377& 0.198\\
 Swin + Transf.& 0.487& 0.167& 0.377& 0.195\\
 Vmamba + LLM & 0.514 & 0.206 & 0.401 & 0.215 \\  \hline
% \toprule [0.5 pt]
\end{tabular}
}
\end{table}

% \begin{figure*}[!h]
\begin{figure*}[!htp]
    \centering
    \includegraphics[width=1\linewidth]{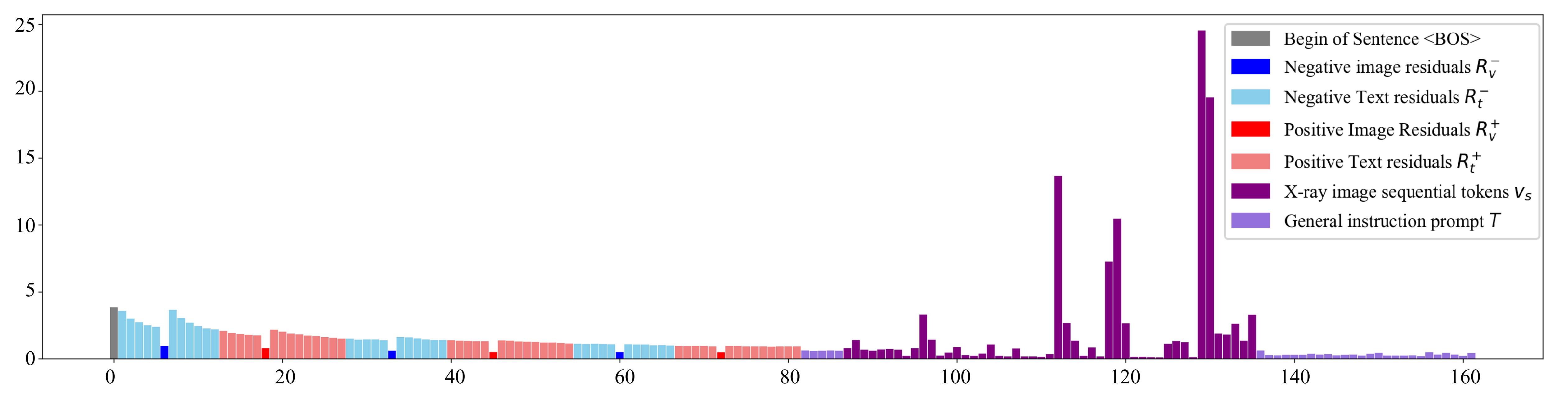}
    \caption{Distribution of aggregated attention weights in the LLM's attention mechanism across token positions for an input X-ray image. Attention scores were computed via column-wise summation across the attention matrix. The x-axis denotes token position indices, while the y-axis represents the computed attention scores. The bar plot demonstrates the predominant influence of image sequential tokens \(v_{s}\), with secondary contributions from negative residuals \(R^{-}\) and positive residuals \(R^{+}\). }
    \label{fig:attention_bar}
\end{figure*}

\begin{figure*}[!htp]
    \centering
    \includegraphics[width=1\linewidth]{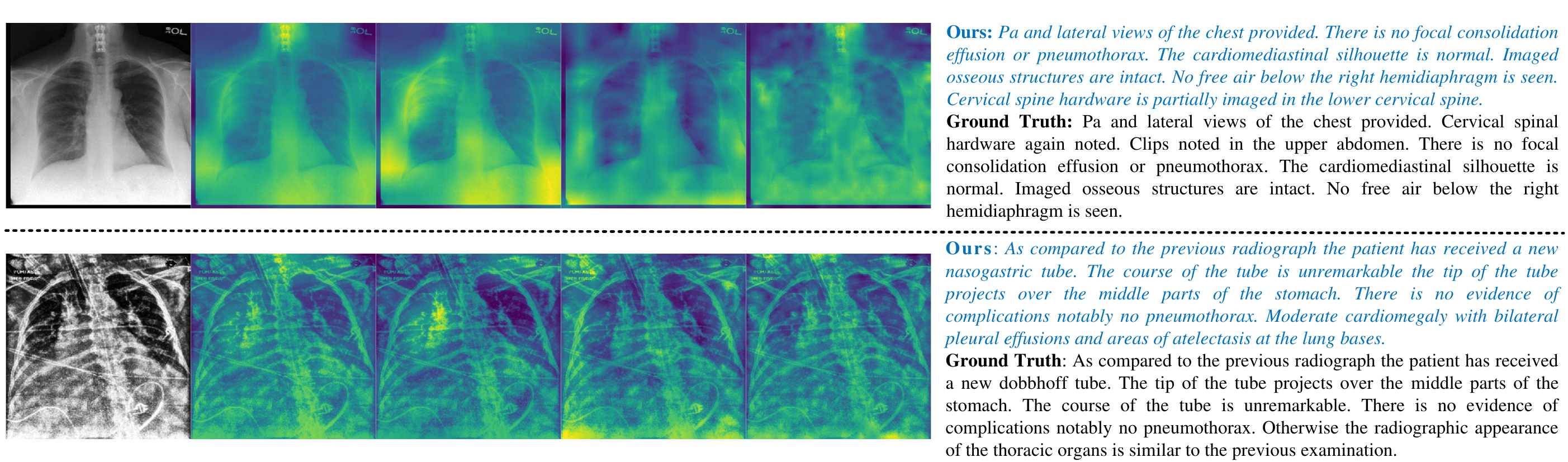}
    \caption{\mmodification{Visualization of the input X-ray image, intermediate feature maps from the Vmamba model blocks, and the generated radiology report on the MIMIC-CXR dataset. }}
    \label{fig:Feat_VIS}
\end{figure*}

\noindent $\bullet$ \textbf{Analysis of Backbone and Decoder Variants.~}
As shown in \tabref~\ref{tab:basc_transformer}, we follow R2Gen~\cite{chen2020R2Gen} and R2Gen-Mamba~\cite{sun2025r2genmamba} by using a lightweight 3-layer, 8-head Transformer decoder instead of a large LLM. This setting provides a fair comparison to earlier baselines and isolates the effect of different visual encoders. We observe that replacing the Swin Transformer with VMamba yields consistent improvements, demonstrating that VMamba contributes to efficiency and accuracy even without relying on a large LLM. Moreover, when paired with a strong LLM (e.g., Qwen1.5-1.8B), our framework achieves further gains, indicating that while the LLM plays a dominant role in language fluency, our proposed retrieval and residual mechanisms contribute complementary benefits beyond backbone strength.
% In \tabref\ref{tab:basc_transformer}, ... we follow R2Gen~\cite{chen2020R2Gen} and R2Gen-Mamba~\cite{sun2025r2genmamba} use 3 layers 8 heads transformer as language decoder instead of LLM. 报告生成的效果大部分归因于LLM能力，我们的方法在此基础上进一步提高了报告生成的能力
}
% [A{'Bleu_1': 0.4934173444360285, 'Bleu_2': 0.32240911811416806, 'Bleu_3': 0.23243008120507733, 'Bleu_4': 0.17573811314493448, 'ROUGE_L': 0.3777015686365582, 'METEOR': 0.19881229127297068, 'CIDEr': 0.4378667333672415}
% [A{'Bleu_1': 0.48717568859551746, 'Bleu_2': 0.30762641455523065, 'Bleu_3': 0.2210578531220855, 'Bleu_4': 0.16694504192801496, 'ROUGE_L': 0.37694579341814405, 'METEOR': 0.19542875993540063, 'CIDEr': 0.41785204393299835}

\subsection{Visualization} 
\noindent $\bullet$ \textbf{X-ray Medical Report.~} 
Fig.~\ref{fig:visual_report} illustrates a side-by-side comparison of X-ray images alongside their respective ground truths and the reports generated by the baseline R2GenGPT model and our model. It is clear that our method is capable of producing reports that closely align with the ground truth, with only minor discrepancies highlighted in mismatching terms but the overall performance of our framework on the MIMIC-CXR dataset is promising. 

% quantifies the attention scores allocated to each token position, with token types differentiated by color according to the legend. The scores are derived through column-wise aggregation of attention weights for a representative X-ray image from the MIMIC-CXR dataset. Analysis reveals that sequential image tokens (\(v_{s}\), purple) receive the highest attention scores, indicating their primary role as carriers of diagnostic information. Residual tokens demonstrate secondary influence, consistent with their auxiliary function in context preservation. Notably, residual tokens attract significantly stronger attention than general instruction prompts (\(T\), medium purple), suggesting their greater contextual relevance to the model's decision-making process.
\noindent $\bullet$ \textbf{Token-wise Attention Distribution.~} 
Fig.~\ref{fig:attention_bar} quantifies the attention scores allocated to each token position, with token types differentiated by color according to the legend. The scores are derived through column-wise aggregation of attention weights for a given X-ray image from the MIMIC-CXR dataset. Sequential image tokens (\(v_{s}\), purple) receive the highest attention scores, indicating their primary role as carriers of diagnostic information. Notably, residual tokens attract significantly stronger attention compared to general instruction prompt (\(T\), medium purple), indicating that the LLM prioritizes these residual contextual prompts during report generation.

\noindent $\bullet$ \textbf{Feature Maps.~} 
The X-ray image and its corresponding feature map are displayed side by side in Fig.~\ref{fig:Feat_VIS}. It is evident from the feature map that the VMamba vision backbone effectively extracts discriminative visual features from the X-ray image, emphasizing regions of interest such as lesions, organs, and other anomalies. These extracted features supply detailed information for subsequent processing.

\subsection{Limitation Analysis}  
Although our proposed R2GenCSR achieves better performance on three large-scale report generation benchmark datasets, however, our model may still limited by the following issues: 
1). We adopt a simple retrieval strategy for context sample mining, more advanced retrieval techniques can be exploited to achieve better performance; 
2). The knowledge about the disease is ignored in our R2GenCSR framework, this information may be useful to guide the X-ray image report generation task. 
In our future works, we will consider improving the proposed framework from the aforementioned two aspects.

\section{Conclusion}  
In this paper, we have developed a novel context-guided efficient radiology report generation framework with the aim of enhancing the performance of Large Language Models in clinical settings. We introduced the Mamba model as a linear complexity vision backbone to effectively extract visual features from X-ray images, ensuring that the computational burden is significantly reduced while maintaining performance parity with the robust Transformer architecture. Our approach was complemented by the integration of vision tokens, context information, and prompt statements to facilitate the generation of high-quality medical reports by the LLM. The proposed framework, extensively evaluated on the \mmodification{IU X-Ray}, MIMIC-CXR, and CheXpert Plus datasets, has demonstrated its efficacy and state-of-the-art performance. Our work not only contributes to the advancement of automated medical report generation but also provides valuable insights for leveraging LLMs in other domains.

\section*{Acknowledgement}
This work was supported by the National Natural Science Foundation of China under Grant 62102205. 
Anhui Provincial Natural Science Foundation-Outstanding Youth Project, 2408085Y032. 
The authors acknowledge the High-performance Computing Platform of Anhui University for providing computing resources.

{
\small
\bibliographystyle{ieeenat_fullname}
\bibliography{reference}
}

\end{document}